# Explainability Through Systematicity: The Hard Systematicity Challenge for Artificial Intelligence

Matthieu Queloz[1]

## Abstract

This paper argues that explainability is only one facet of a broader ideal that shapes our expectations towards artificial intelligence (AI). Fundamentally, the issue is to what extent AI exhibits *systematicity*—not merely in being sensitive to how thoughts are composed of recombinable constituents, but in striving towards an integrated body of thought that is consistent, coherent, comprehensive, and parsimoniously principled. This richer conception of systematicity has been obscured by the long shadow of the "systematicity challenge" to connectionism, according to which network architectures are fundamentally at odds with what Fodor and colleagues termed "the systematicity of thought." I offer a conceptual framework for thinking about "the systematicity of thought" that distinguishes four senses of the phrase. I use these distinctions to defuse the perceived tension between systematicity and connectionism and show that the conception of systematicity that historically shaped our sense of what makes thought rational, authoritative, and scientific is more demanding than the Fodorian notion. To determine whether we have reason to hold AI models to this ideal of systematicity, I then argue, we must look to the *rationales* for systematization and explore to what extent they *transfer* to AI models. I identify five such rationales and apply them to AI. This brings into view the "hard systematicity challenge." However, the demand for systematization itself needs to be regulated by the rationales for systematization. This yields a *dynamic* understanding of the need to systematize thought, which tells us *how* systematic we need AI models to be and *when*.

---

✉ Matthieu Queloz
matthieu.queloz@unibe.ch

1 Institute of Philosophy, University of Bern, Laenggassstrasse 49a, 3012 Bern, Switzerland

# 1 Introduction

Concerns about recent advances in artificial intelligence (AI) are often articulated in terms of a perceived lack of *interpretability* or *explainability*.[1] The definitions of these terms vary, sometimes centering on the interpretability or explainability of the *models* themselves, sometimes on the interpretability or explainability of the *outputs* produced by these models.[2] Nevertheless, the same underlying difficulty lies at the root of both lines of inquiry: deep artificial neural networks operate via functions of such extreme complexity that their input–output mappings resist straightforward comprehension. This renders the inner workings of the models mechanistically opaque and gives rise to *model-centric* demands for interpretability and explainability.[3] But it concurrently creates a different and often more immediately challenging form of opacity, namely rational opacity concerning how a given output was derived and why we should accept it. This second form of opacity fuels *output-centric* demands for interpretability and explainability.

My aim is to propose a conceptual reframing of output-centric demands for interpretability and explainability that can unify and deepen our understanding of them. I submit that the demands for the interpretability and explainability of outputs are but facets of an older and broader ideal shaping our expectations towards artificial intelligence, and that reframing these demands in terms of that broader ideal can place and connect these demands while also bringing into view what they leave out.

On the reframing I propose, the fundamental question, especially from the perspective of users as opposed to builders of these models, is to what extent AI exhibits *systematicity of thought* in the sense in which eighteenth-century theorists of cognition like Christian Wolff, Immanuel Kant, and Johann Heinrich Lambert used the phrase, i.e. to what extent model outputs can be integrated into a body of thought that is consistent, coherent, comprehensive, and parsimoniously principled. This is a notably wider, richer, and more demanding sense of the phrase "the systematicity of thought" than the one that has dominated cognitive science and AI research for the past three decades.

In AI research, the question of whether a model exhibits "systematicity of thought" is sometimes treated simply as a matter of whether it is capable of handling *compositionality*, i.e. whether it is sensitive to how thoughts are *structured* in the sense of being *composed of recombinable constituents*.[4] This use of the phrase goes back to Jerry Fodor, who, in a string of papers with various colleagues, suggested that network architectures were fundamentally at odds with "the systematicity of thought" in this narrow sense. He thereby laid down what came to be known as the "systematicity challenge" to neural networks.[5]

---

[1] On the history of demands for interpretability and explainability in AI research, see Confalonieri et al. (2021). For overviews of various attempts to meet them, see Gilpin et al. (2018), Minh et al. (2022), and especially Hsieh et al. (2024), which offers a book-length survey of current work on explainable AI.

[2] See Kim et al. (2018); Benchekroun et al. (2020); Krishnan (2020); Buijsman (2022); Duque Anton et al. (2022); Buchholz (2023); Rizzo et al. (2023); Nanda (2024); Räz (2024b).

[3] See Mann et al. (2023). This has prompted a research programme of its own—"mechanistic interpretability"—which seeks to elucidate the inner workings of those networks; see Elhage et al. (2021) for a seminal example. For philosophical discussions of the opacity of neural networks, see Räz and Beisbart (2024) and Beisbart (forthcoming).

[4] Lake and Baroni (2023) offer a prominent example of this tendency. But this involves a simplification that is worth resisting, as Spenader and Blutner (2007) have argued.

As we shall see, however, that challenge has now lost much of its challengingness. The time has therefore come to shift our attention to what might be called the "hard systematicity challenge": the challenge of building neural networks that are sensitive to the structure of thought not merely in the narrow sense of being sensitive to how thoughts are composed of recombinable constituents, but in the broader and more demanding sense of striving towards an integrated body of thought that is consistent, coherent, comprehensive, and parsimoniously principled. This involves stepping back from the inner articulation of individual thoughts to consider how an entire array of thoughts forms an orderly structure—it marks what I shall refer to as a shift from *micro-* to *macrosystematicity*.

Ever since Archimedes' systematization of statics and Euclid's systematization of geometry provided lasting paradigms of systematic thought, this more demanding notion of systematicity has been deeply influential in the development of Western thought and science, shaping conceptions of intelligence by informing people's sense of what it means for thought to be rational, authoritative, and scientific (Gaukroger, 2020; Hoyningen-Huene, 2013; Rescher, 1979, 1981, 2000, 2005). By the eighteenth century, one historian notes, systematicity was the "unquestioned answer to the question of the nature of reason's fundamental demand" (Franks, 2005, p. 3). I argue that this ideal of systematicity is not only fundamental to making the outputs of AI models interpretable to begin with, which is the first step towards explainability, but also underlies our sense of what makes for *good* explanations. We do not just need explainability—we need explainability *through* systematicity.

This comes out clearly when considering large language models (LLMs). With the advent of transformer-based architectures (Vaswani et al., 2017), LLMs have become remarkably proficient at generating linguistic outputs that are grammatically correct and contextually relevant; if suitably prompted, they can also generate explanations to accompany those outputs. But the remaining limitations of LLMs reveal themselves precisely when these explanations flout the ideal of systematicity: when an explanation turns out to be inconsistent with the facts or with other outputs of the model, or when it feels rationally disconnected and thus fails to cohere with them; in certain cases, the explanation may also feel insufficiently principled, or insufficiently parsimonious in the number of principles it invokes—what we might call the "If you don't like my principles, I have others"-problem. What is lacking in all those cases is not interpretability or explainability. Nor is it systematicity in the narrow, Fodorian sense at issue in the "systematicity challenge." What is lacking is systematicity in the broader, more demanding sense of macrosystematicity. And the goal of this paper is to recover this more demanding notion of systematicity from under the shadow of the systematicity challenge and show how this notion can inform our sense of what we need AI models to be like.

---

[5] See Calvo and Symons (2014); Verdejo (2015).

Before we can fully appreciate this richer notion of systematicity, however, it is necessary to address the influential "systematicity challenge" posed by Fodor and his colleagues (§2). To clear the conceptual ground for an alternative conception of systematicity, but also to develop the conceptual building blocks for the paper's positive contribution and to situate that contribution in relation to the prevailing paradigm, I shall engage with Fodor's arguments in some detail (§§3–5). In particular, I shall bring out how the Fodorian inferences from the "systematicity of thought" to the language-like and atomistic character of thought can be resisted. This serves four purposes within the paper: first, to decouple the notion of systematicity from the symbolic and atomistic architecture with which Fodor so influentially associated it; second, to defuse the perceived tension between systematicity and network architectures; third, to distinguish four different senses of the phrase "the systematicity of thought"; and finally, to determine how Fodor's influential but narrow notion relates to the broader and more demanding ideal of systematicity that shaped our sense of what makes thought rational, authoritative, and scientific.

To determine whether we have reason to hold AI models to this more demanding ideal of systematicity, I then argue, we must look to the *rationales* for systematizing thought and explore to what extent they *transfer* to AI models (§6). Drawing on the history of efforts to systematize human thought, I recover five such rationales. This brings into view the "hard systematicity challenge" (§7). As I argue in closing, however, the demand for systematization itself needs to be regulated by the rationales for systematization. This yields a *dynamic* understanding of the need to systematize thought, which draws on a sense of why AI models need to be systematic to tell us *how* systematic we need AI models to be and *when* (§8).

## 2 The Long Shadow of the Systematicity Challenge

If there is a notion of systematicity that has historically been influential in shaping our conceptions of rational, authoritative, and scientific thought, why has it not received more attention in the literature on artificial intelligence? A likely explanation is that this traditional notion has been overshadowed by the notion of systematicity at work in the "systematicity debate" that loomed large over the past three decades, hogging much of the attention around systematicity to this day. The debate was initiated by Jerry Fodor and Zenon Pylyshin (1988), who influentially harnessed the phrase "the systematicity of thought" to challenge the view that artificial neural networks offer good models of cognitive processes.

In what became known as the "systematicity challenge" (Calvo & Symons, 2014; Verdejo, 2015), Fodor and Pylyshin pointed out that there is a revealing *symmetry* in our cognitive capacities: the ability to entertain a given thought, such as "John loves Mary," implies the ability to entertain certain other thoughts, such as "Mary loves John." This symmetry has been thought to be a key criterion of concept possession, including in AI (Butlin, 2023, p. §3). Fodor and Pylyshin referred to this phenomenon as "the systematicity of thought," and contended that network architectures are inherently ill-suited to account for it—in particular, to account for it *as a matter of nomic necessity*, as Fodor and McLaughlin (1990) clarified.[6]

Here, the *classical* computational theory of mind, which understands cognition as rule-governed symbol manipulation running on a classical, Turing-style computational architecture (consisting of a central processor and a memory on which to inscribe a string of symbols), clearly has the edge over neural network architectures, Fodor and his co-authors thought. This was because the classical theory postulates that thought itself consists of discrete symbols that can be combined, permuted, and substituted by symbols of the same kind, thereby virtually *guaranteeing* the emergence of the phenomenon they dubbed "the systematicity of thought."

By contrast, the *connectionist* theory, which proposes to model cognition using neural networks, looks less well-placed to account for "the systematicity of thought." Neural networks encode information not in strings of discrete symbols, but sub-symbolically, in the activation thresholds or "biases" of nodes and the connection strengths or "weights" between these nodes. Whereas classical models serially process symbols that are discrete and categorical (something is either a "cat" or "not a cat"), connectionist models dispense with symbols altogether, relying instead on the parallel processing of sub-symbolic activation patterns that are distributed across the network and continuous-valued (something is "cat-like" to a certain degree).[7] To connectionists like Geoffrey Hinton, symbols are the "luminiferous aether of AI" (Russell & Norvig, 2021, p. 42)—comparable to the mythical medium that nineteenth-century physicists postulated to account for the propagation of electromagnetic waves, but eventually found they could dispense with altogether.

The systematicity debate between classicalists and connectionists was never about whether a connectionist architecture was compatible *in principle* with information processing exhibiting the symmetry that Fodor and Pylyshin labeled "the systematicity of thought"—it can be mathematically demonstrated that a classical symbol processor can, in principle, be implemented within a connectionist architecture.[8] Rather, the debate was over whether a connectionist architecture is well-suited to *account for* the systematicity of thought, and whether it would have to implement symbolic processing to reliably reproduce human-like systematicity.[9]

---

[6] As others have noted, however, explaining systematicity as necessitated by laws is a very strong requirement that classical architectures likely cannot meet either (Buckner and Garson 2019).

[7] Though the advent of hybrid architectures such as those described in Smolensky (1990) and Eliasmith (2013) has rendered the distinction between symbolic and sub-symbolic processing less clear-cut. The two can in principle be combined to varying degrees.

[8] Building on seminal work by McCulloch and Pitts (1943), Siegelmann and Sontag (1992) proved that RNNs are Turing-complete given unbounded precision, memory, and computation time; see Weiss et al. (2018) for RNNs under more realistic constraints, and see Pérez et al. (2021) for transformer-based networks.

[9] For classicalist positions, see Fodor and McLaughlin (1990); McLaughlin (1993); Aydede (1997); McLaughlin (2009); for connectionist positions, see Smolensky (1988, 1990); van Gelder (1990); Matthews (1994); Cummins (1996); Cummins et al. (2001). For thorough accounts of the debate, see Aizawa (2003), the essays in Calvo and Symons (2014) as well as Buckner and Garson (2019) and Rescorla (2024), and see Johnson (2004) for a critique of the conception of systematicity that is supposed to underpin it. Recently, "category theory," which draws on the mathematical theory of structure, was put forward as a third contender in the debate; see Phillips and Wilson (2016).

The long shadow of the systematicity challenge helps explain why older, broader, and more demanding notions of systematicity have largely remained invisible. Fodor and Pylyshin succeeded in framing the debate over connectionism for decades, focusing connectionist efforts on overcoming their challenge.[10] The relevant notion of systematicity was crystallized into benchmarks that connectionist models then sought to pass (Ettinger et al., 2018; Hadley, 1994; Yu & Ettinger, 2020). By thus turning a spotlight on this particular conception of systematicity, AI research effectively cast older conceptions into the shadows.

Furthermore, this systematicity challenge turned out to be hard enough already. Even neural networks trained on vast amounts of data have struggled with forms of systematicity that symbolic models handle with ease (Hupkes et al., 2020; Marcus, 2001; Press et al., 2023). This implicitly reinforced the idea that there was little point worrying about broader and more demanding forms of systematicity as long as neural networks had difficulties reliably reproducing the systematicity of thought in Fodor's comparatively basic sense.

## 3 Beyond the Systematicity Challenge

Yet recent advances have made the systematicity challenge lose much of its challengingness. Neurocompositional computing has enabled connectionist models to make great strides (Smolensky et al., 2022), and in 2023, a *Nature* article triumphantly announced that techniques such as meta-learning for compositionality (MLC), which involves using metadata from previous learning episodes to improve the learning process (thereby allowing the networks to "learn to learn"), enable neural networks to rise to Fodor and Pylyshin's challenge (Lake & Baroni, 2023). This has been hailed as a "breakthrough in the ability to train networks to be systematic" (Smolensky, quoted in Kozlov & Biever, 2023, p. 16). As Raphaël Millière and Cameron Buckner observe, "the remarkable progress of LLMs in recent years calls for a reexamination of old assumptions" about systematicity as a "core limitation of connectionist models" (2024, §3.1).[11]

However, the systematicity challenge is more than an empirical prediction that neural networks will struggle to replicate the systematicity of human cognition. It is also a set of theoretical arguments, which Fodor developed across numerous publications (Fodor, 1998, 2000, 2003; Fodor & Lepore, 2002). On the resulting view, the systematicity of thought constitutes evidence that the substrate of thought must be language-like rather than network-like in structure. Fodor encourages us to infer that the observed symmetry in our cognitive capacities is best explained by postulating a *language of thought*: an inner, private language, "Mentalese," which involves

---

[10] See Hadley (1997, 2004) and Buckner and Garson (2019) for critical evaluations of some of these attempts.

[11] As Quilty-Dunn et al. (2023) argue, however, the symbolic approach to computational cognitive science lives on nonetheless, notably in approaches invoking a "probabilistic language of thought" (Colombo 2023; Goodman et al., 2015).

the rule-governed manipulation of discrete and partly innate mental representations or symbols.

In addition to this psychological hypothesis, Fodor also takes the systematicity of thought to support a semantic hypothesis: the best explanation for the systematicity of thought is that thought is fundamentally atomistic, i.e. made up of unstructured concepts or primitive entries in our mental lexicon that derive their meaning or content solely from their relation to entities in the environment rather than from their relation to other concepts. Furthermore, these atomistic constituents must be recombinable according to syntactic rules, forming complex thoughts whose meaning or content is simply a function of the concepts they are composed of along with the way they are syntactically combined. In other words, Fodor takes the systematicity of thought to reflect its *atomism*, *combinatorial syntax*, and *compositional semantics*. Together, these three features also account for the *productivity* of thought. Only a finite number of constituents need to be learned to generate an infinite variety of new thoughts—we can, in Humboldt's phrase, make "infinite use of finite means"—because these new thoughts are generated by recombining familiar constituents, producing unprecedented thoughts that can nonetheless be understood thanks to the compositionality of their semantics.

In sum, Fodor takes the systematicity of thought to count against a connectionist view of cognition and in favor of a classical view because that systematicity is best explained by (a) the psychological thesis that thought is language-like in structure, involving the rule-governed manipulation of symbols, and (b) the semantic thesis that meaning is atomistic and compositional in nature.

To overcome the systematicity challenge, it is therefore not sufficient to ace benchmarks. What one takes empirical results to show depends on one's theory-laden interpretation of the results, and Fodor takes the systematicity of thought to impose a certain theoretical view of cognition on us. To overcome the systematicity challenge at the theoretical level, one needs to understand how the crucial inferences that Fodor makes can be resisted.

Accordingly, I propose to indicate how connectionists can resist these Fodorian inferences, thereby clearing the path for a way of thinking about systematicity in neural networks that aspires to a more demanding form of systematicity. Moreover, doing so will give us occasion to draw an important distinction that will help us make sense of how Fodor's narrow notion of systematicity relates to broader notions of systematicity.

### 3.1 Resisting the Inference from Systematicity to the Language of Thought

To resist the inference from the systematicity of thought to the language of thought, we must first recognize that there is a significant ambiguity in the phrase "the systematicity of thought." It is ambiguous between the *activity* of thinking and the *object* of that activity, namely *what is thought*:

- *The systematicity of thinking*: on the one hand, the "systematicity of thought" can refer to the systematicity of *thinking*. This refers to the patterns, intercon-nections, and regularities discernible in the activity of thinking as performed by actual thinkers: which thoughts they can entertain, whether the capacity to entertain some thoughts entails or tends to entrain the capacity to think others, and whether the capacity to draw certain inferences entails or tends to entrain the capacity to draw other inferences. The pattern that Fodor and colleagues focus on is a particular *symmetry* observable in human cognitive capacities: thinkers capable of entertaining thoughts of the form *aRb* also have the capacity to entertain thoughts of the form *bRa*. Fodor regards this symmetry in human thinking as a contingent empirical fact (1998, 26).

- *The systematicity of what is thought:* on the other hand, the "systematicity of thought" can also refer to the systematicity of *what is thought*. This is the idea that there is a *structure* inherent in the contents of thought. Typically, this is taken to mean that thoughts are *articulated in terms of recombinable constituents*.[12] The possibilities for well-formed recombinations are not random, but follow rule-governed patterns that in turn give rise to systematic interconnections between thinkable contents. This is a claim about the nature of thought itself, not an empirical observation about thinkers. Each actual thought is situated in a structured network of thinkable contents. This structure is logically independent of the capacities actual thinkers in fact possess.

This ambiguity in the phrase "the systematicity of thought" is an instance of a more widespread phenomenon that has been called *the act-object ambiguity* (Alvarez, 2010, p. 125). Once we recognize this ambiguity, we can see that Fodor's argument from the "systematicity of thought" is really an argument from the systematicity of *thinking*. He observes that, *as a matter of empirical fact*, thinkers capable of entertaining thoughts of the form *aRb* also have the capacity to entertain thoughts of the form *bRa*. This is not meant to be an a priori, conceptual point, but a contingent, empirical one. Fodor takes it to be "conceptually possible that there should be a mind that is able to grasp the proposition that Mary loves John but not able to grasp the proposition that John loves Mary. But, in point of empirical fact, it appears that there are no such minds" (1998, p. 26). Note also that the relevant form of systematicity "concerns symmetries of cognitive *capacities*, not of actual mental states" (1998, 26n2). Not everyone who thinks that "Humans walk dogs" also thinks that "Dogs walk humans." The point is merely that if they *can* think the former, they *can also* think the latter. Likewise, Fodor observes that thinkers capable of inferring "*p*" from "*p and q*" are also capable of inferring "*m*" from "*m and n*." Again, however, this is, in the first instance, a pattern in our *thinking*. Our inferential capacities, like our capacities to entertain certain thoughts, come in bundles, so that whoever possesses one capacity also possesses the other. Yet this is in the first instance an observation about the constant conjunction of certain skills, not one about the inherent structure of what is thought.

---

[12] This sense of systematicity figures centrally in Peacocke (1992, 42), Cummins (1996, 2010), Johnson (2004), Perler (2004), and Salje (2019).

In a second step, Fodor argues that the observable systematicity of thinking requires explanation. Why is it that certain cognitive capacities go together in this way? His answer is that the systematicity of thinking must reflect a systematicity in thought itself: the thinkable contents must themselves be articulated in terms of recombinable constituents. So far, so uncontroversial.

But Fodor then makes a crucial further inference: he takes the systematicity of thinking as evidence for the reality of a private mental language: the language of thought, which he imagines as consisting of symbols or discrete mental representations with syntactic and semantic properties closely resembling those of natural languages.

Fodor's argument is therefore not really an argument *from* the systematicity of thought; it is better described as an argument from the systematicity of thinking *to* the systematicity of thought—where the latter is taken to refer to the combinatorial syntax and compositional semantics of a private language of thought.[13]

Notice, however, that this argument draws part of its plausibility from the fact that it trades on a further equivocation: it is *prima facie* compelling to say that the systematicity of *thinking* is based on the underlying systematicity of *thought*, and that thought must therefore have some inherent structure; but the familiar phenomenon we call "thought," whose public manifestations we encounter in everyday linguistic and non-linguistic behavior, and which is widely agreed to be analyzable into discriminable—though not necessarily detachable (Ryle, 2009, p. 192)—concepts, is one thing; the inner, private language of thought that Fodor invites us to postulate is quite another. Even after we have disambiguated thinking from thought, therefore, it seems that the systematicity of thought still *comes in twice* in Fodor's story: an inner, private representational system is postulated to explain the systematicity of thinking, which in turn explains the systematicity of publicly manifestable thought. By "publicly manifestable thought," I mean thought content (e.g., the proposition expressed by "Mary loves John") whose existence, structure, and attribution to an agent are grounded in publicly observable evidence, such as linguistic utterances and non-linguistic actions. This is the common object of psychological explanation and interpretation—the beliefs, desires, intentions we attribute based on behavior. It stands in contrast to Fodor's notion of an essentially *private* "language of thought," which is postulated as an underlying, unobservable causal mechanism.

Yet, from the fact that thought—the publicly manifestable product of our thinking capacities—is articulated in terms of recombinable constituents, it simply does not follow that we must each think in terms of a private system of mental symbols. This inference is explicitly blocked by none other than Gareth Evans, one of the most influential proponents of the idea that "thoughts are *structured*" (1982, p. 100). To make sense of this structure, Evans invites us to see the thought that $a$ is $F$ "as lying at the intersection of two series of thoughts: the thoughts that $a$ is $F$, that $a$ is $G$, that $a$ is $H$, …, on the one hand, and the thoughts that $a$ is $F$, that $b$ is $F$, that $c$ is $F$, …, on the other" (1982, p. 209). As Ludwig Wittgenstein remarked to Friedrich Waismann, it is these possibilities of permutation and substitution that give point to the structure of thought (Waismann, 1979, p. 90). Were these systematic variations not in the offing, thought might as well have no internal structure at all. From this observation, Evans derives what he calls the *generality constraint*: "if a subject can be credited with the thought that *a* is *F*, then he must have the conceptual resources for entertaining the thought that *a* is *G*, for every property of being *G* of which he has a conception" (1982, p. 104).[14] Evans thus agrees with Fodor that there is a structure inherent in well-formed thoughts that allows systematic variants of them to be produced and understood, where systematic variation involves permuting constituents or, more demandingly, substituting constituents of the same kind.

---

[13] This comes out clearly in Fodor and Pylyshin (1988, 26n25).

But Evans is careful to block the inference from the systematicity of thought to the language of thought hypothesis. He remarks, right after asserting that thoughts are structured: "This might seem to lead immediately to the idea of a language of thought … However, I certainly do not wish to be committed to the idea that having thoughts involves the subject's using, manipulating, or apprehending symbols" (1982, pp. 100–101). There is a danger here of overintellectualizing human thought and modeling the structure of sub-personal processing too closely on the structure of natural language. We should not be too quick to infer the structure of a cause from the structure of its effect (Churchland & Sejnowski, 1990). Maybe part of what makes natural language so useful is precisely that it introduces a structure we would otherwise lack.

Instead of seeing the structure of language as mirroring a pre-existing structure in thought, we can explain the sense in which thoughts are structured "in terms of their being a complex of the exercise of several distinct conceptual *abilities*," Evans suggests; "someone who thinks that John is happy and that Harry is happy exercises on two occasions the conceptual ability which we call 'possessing the concept of happiness'" (1982, p. 101). Evans's perspective implies that the fundamental units of our cognitive life are our conceptual abilities—the ability to identify and reidentify things that are F as opposed to non-F, to classify these things as being F, and to know what their being F implies.[15] To understand why our conceptual abilities exhibit certain systematic patterns, such as the symmetry observed by Fodor and Pylyshin, we need to look at how conceptual abilities are acquired and how their formation and repeated exercise affects and interacts with the formation and repeated exercise of other conceptual abilities. These conceptual abilities are not the externalization of some innate inner blueprint. Rather, our thought is the internalization of the thinking techniques we acquire through enculturation and acculturation.

---

[14] For a defense of an unrestricted version of this constraint, see Camp (2004); for a defense of a weaker version of the constraint, see Dickie (2010); and see Travis (2015) for a critique of it that echoes Warren Goldfarb's Wittgensteinian gripes about the supposed "fixity of meaning" (1997).

[15] This need not amount to an *identification* of concepts with abilities. For the reasons enumerated in Glock (2006, 2009a, b, 2010a, b, 2020), this would be too simple; concepts occur or are involved in thoughts in a way in which abilities are decidedly not. Nonetheless, to possess a concept is to possess certain discriminatory, classificatory, and inferential abilities. That much is granted even by Fodor: "*having* a concept is: *being able* to mentally represent (hence to think about) whatever it's the concept of" (2003, 19, emphasis added).

Thus, we can resist Fodor's inference from the systematicity of thinking to the systematicity of thought by realizing that the direction of explanation can just as well be reversed. Perhaps it is not that the systematicity of thinking reflects the antecedent systematicity of thought. Perhaps the systematicity of thought is the *product* of the systematicity of thinking. Thought would then be systematic because systematic patterns—such as certain symmetries—dynamically emerge from the way we form and exercise conceptual abilities. Accordingly, there is no need to postulate a language of thought to explain systematicity in our thinking.

### 3.2 Resisting the Inference from Systematicity to Atomistic and Compositional Semantics

Fodor also argues that the systematicity in our thinking points not just to a symbolic structure, but to a semantically atomistic and compositional one, whereby the meaning of any given thought can be fully understood in terms of the way it combines discrete and unstructured atomic concepts whose meaning is fixed, insensitive to context, and independent of other concepts (1998). He presents this as a challenge to connectionist models, which lack discrete, atomic elements due to their reliance on distributed, continuous-valued activation patterns. In a neural network, meaning is distributed across many nodes, and each node's activation is modulated by its weighted connections to other nodes. This interdependency means that the semantic content of any given pattern cannot be fully understood in isolation. Connectionists are therefore committed to a form of *holism* about meaning. As a result, they cannot simply appeal to the atomistic and compositional nature of meaning to account for the systematicity of thinking, because their semantic holism contradicts the idea that meaning is atomistic and compositional.

Yet connectionists can resist Fodor's inference from systematicity to atomistic and compositional semantics. The key, as Brandom (2008, pp. 134–36) points out against Fodor and Lepore (2002), is to distinguish the *compositionality of meaning* from the *recursiveness of thinking*.[16] It is one thing to say that thoughts can be recursively constructed, in that simpler conceptual building blocks can be endlessly combined and embedded within more complex constructions; but it is quite another thing to say, with Fodor, that the meaning of complex thoughts is fully *interpretable in terms of* their simpler constituents. For notice that on a holistic view, the meaning of complex thoughts can be fully *determined by* the meaning of their simpler constituents *without being fully interpretable in terms of* those simpler constituents. That is, one cannot compute the semantic value of a complex thought without considering its relations to other thoughts—which thoughts it is compatible or incompatible with, and which implications of other thoughts it is compatible or incompatible with. But this non-compositional, holistic semantics at the level of complex thoughts is compatible with full recursiveness *between* levels of constructional complexity. It is this recursiveness that is really needed to ensure the systematicity (and productivity) of thinking. And this recursiveness does not require atomistic and compositional semantics. Rather, recursive rules of combination can operate alongside a semantic holism wherein context and inferential connections further refine or modify meaning through a recursive layering of inferential connections that dynamically shape meaning at each level of constructional complexity.[17] This layering means that while complex thoughts are built recursively from simpler constituents, the full inferential significance of any constituent is not fixed atomically but is modulated and potentially transformed by the larger constructions it enters into. Each level of embedding potentially brings new inferential relations into play, refining the meaning beyond a simple compositional sum of pre-existing parts.

---

[16] For a critique of this line of thought, see Fermüller (2010); for a corresponding defence, see Turbanti (2017, ch. 4, §4.2).

Connectionists can accordingly reject the idea that the meaning of a complex thought can be understood wholly bottom-up, independently of its relation to other thoughts, while also maintaining that our thinking is recursive in allowing new thoughts to be formed from simpler conceptual building blocks.

Far from precluding the systematic co-occurrence of cognitive capacities that Fodor points to, in fact, semantic holism can account for this co-occurrence not just as an empirical regularity, but as reflecting the holistic character of thinking itself: it is constitutive of genuinely *thinking* a thought that one can place it in a web of systematic relationships to other thoughts. There could not be a thinker capable only of entertaining a single thought (Clark, 1991).

The same holistic point also applies to the possession of individual concepts. While Fodor's atomistic account expressly "denies that the grasp of *any* interconceptual relations is constitutive of concept possession" (Fodor, 1998, p. 71), holistic accounts such as those of Quine (1951), Sellars (1958), or Brandom (1994) maintain that grasping at least some of the inferential connections a concept stands in to other concepts *is* constitutive of concept possession. In Brandom's slogan, "one must have many concepts in order to have any" (1994, p. 89).[18] To genuinely count as grasping a concept, one has to grasp its role in a web of inferential connections between thoughts, for these are a crucial part of what gives determinate content to the concept in the first place.[19] Concepts would be mere labels, devoid of substantive meaning, if conceptual content were limited to the referential dimension a concept bears to its object. Imagine being handed an *F*-detector that lights up when and only when presented with an *F*; you could use it to sort things into *F*s and non-*F*s, but as long as you lacked any idea of what something's being *F implied*, you would not understand the *significance* of being *F* or non-*F*; it would be an empty label to you.[20] As Wilfrid Sellars puts the point: "It is only because the expressions in terms of which we describe objects … locate these objects in a space of implications, that they describe at all, rather than merely label" (1958, 306–7).

---

[17] For an account of meaning in neural networks along those lines, see Piantadosi (2021), who argues that neural networks could acquire concepts by adjusting their parameters so that they church-encode conceptual roles. See also Piantadosi and Hill (2022); Piantadosi et al. (2024).

[18] See also Brandom (2009, 202; 2019, 113).

[19] Whether this constitutes a necessary or a sufficient part of the determination of conceptual content depends on the strength of the inferentialism one endorses; Brandom endorses *strong* inferentialism, on which inferential articulation, broadly construed (i.e. including materially correct inferences as well as noninferential circumstances and consequences of concept application), is *sufficient* to account for conceptual content (2000, 28; 2007, 657).

The systematicity of thinking is thus not a decisive argument against semantic holism, and therefore does not constitute an insuperable theoretical hurdle for connectionism either. With these empirical and theoretical hurdles for connectionism now cleared, it is time to recover, from under the shadow of the systematicity challenge, an older, broader, and more demanding conception of systematicity. Having shown how Fodor's key inferences from systematicity to the language-like and atomistic character of thought can be resisted, we have not only cleared the ground for alternative approaches to cognition, but also begun to devise the conceptual tools that are essential for recovering and articulating the richer, pre-Fodorian conception of systematicity to which we now turn.

## 4 Micro- vs. Macrosystematicity

Our engagement with Fodor's arguments has brought out how "the systematicity of thought" was ambiguous between the systematicity of *thinking*, which refers to the idea that there is a symmetrical pattern in the distribution of human cognitive capacities, and the systematicity of *what is thought*, which refers to the idea that there is a *structure* inherent in thought.

However, we now also need to disambiguate between two ways in which *what is thought* might be said to be systematic or structured:

- *Microsystematicity*: on the one hand, what is thought, i.e. propositions, might be said to be "systematic" or "structured" in the sense of possessing an *inner articulation* or *microstructure*. This is the structure inherent in thought that Wittgenstein, Evans, or Fodor refer to, and they think of it primarily as a *binary* or *all-or-nothing* property. They do not ask *how* systematic or structured thinkable contents are. The point is rather that there is *some* structure as opposed to none.
- *Macrosystematicity*: on the other hand, what is thought, i.e. propositions, might be said to be "systematic" or "structured" in the sense that an entire array of explicitly formulated propositions might fit together into a consistent and coherent whole. This goes beyond the inner articulation or microstructure of thinkable contents. It concerns their *macrostructure*, i.e. the extent to which different propositions can be assembled into an integrated system. This is a *gradable* rather than a binary property, which an array of propositions might exhibit to a greater or lesser degree.

Macrosystematicity requires more than being structured in the minimal sense of being articulated in terms of recombinable constituents. It requires being structured *in a particular way*—namely so as to embody order and harmony. There are multiple aspects to this, and various accounts of systematicity have been proposed, some articulating nine (Hoyningen-Huene, 2013, pp. 35–36) or even eleven dimensions of systematicity (Rescher, 1979, pp. 10–11, 2005, pp. 25–26). But many of these dimensions are either domain-specific or else subsumable under the following five principal dimensions of systematicity:

1. *Consistency*: absence of contradictions.
2. *Coherence*: interconnectedness through relations of rational support.
3. *Comprehensiveness*: avoidance of lacunae.
4. *Principledness*: subsumption of the particular under general principles or laws.
5. *Parsimony*: economy of laws or principles.

---

[20] For a similar illustration, see Brandom (2009, 202).

This 3C2P conception of macrosystematicity, as we might quasi-acronymically call it, zooms out to consider how an entire set of thoughts can be systematic in virtue of how the thoughts relate to one another. While there is a minimal sense in which the inner articulation of thought in terms of recombinable constituents already produces systematic interconnections between thinkable contents (think of Evans's "intersections"), this invokes a sense of "systematic" that contrasts primarily with randomness or arbitrariness: the combinations are not random, but follow patterns. The five dimensions highlighted by the 3C2P conception of macrosystematicity, however, point to forms of systematic *integration and order* that contrast primarily with *fragmentation and disorder*.

Being sensitive to these dimensions allows assessments of the degree to which a given set of propositions instantiates systematic integration and order: consistency is a matter of avoiding contradictions that undermine the integrity of the systematic structure, while coherence more positively strengthens this integrity by establishing rational interconnections among propositions such that they form a network in which each proposition rationally supports or is supported by another. Comprehensiveness further extends systematicity by ensuring that the set covers all the relevant aspects within its scope, without salient gaps or arbitrary omissions. Principledness, in turn, introduces a hierarchical dimension by organizing particular claims under general principles or laws, thereby making the different parts of the structure more unified. Finally, since principles cannot do much unifying if there are as many principles as there are particulars, scoring highly along the dimension of parsimony conforms to the idea that principles or laws should not needlessly proliferate, but should become fewer in number the more general they get.

The 3C2P conception of macrosystematicity is gradable because a set of propositions can exhibit greater or lesser degrees of consistency, coherence, comprehensiveness, principledness, and parsimony, depending on the extent to which it avoids contradictions, establishes relations of rational support, eliminates lacunae, subordinates particulars to principles, and displays economy of principles. Accordingly, this conception of macrosystematicity defines a spectrum along which sets of propositions can be evaluated in a nuanced way, with more systematic sets being more harmonious, organized, and unified.

## 5 Macrosystematicity as an Ideal Regulating Thinking

In relation to the earlier divide between the systematicity of *thinking* and the systematicity of *what is thought*, the 3C2P conception of macrosystematicity is firmly on the side of what is thought. It is a descriptive metric for the systematicity of a given array of propositions. The final step we now need to make, therefore, is to carry back this richer understanding of how thought can be structured to where we started from, which was on the side of *thinking*. If Fodor and Pylyshin's "systematicity of thought" referred to patterns in the practice of thinking which led them to highlight a corresponding form of microsystematicity in the structure of what is thought, how can the recovery of a form of macrosystematicity in the structure of what is thought inform our understanding of the practice of thinking?

When applied to thinking rather than to the structure of what is thought, macrosystematicity refers to an *ideal* governing what the practice of thinking should be like: the ideal whereby thinking *should strive for* systematicity. Instead of being *descriptive* of patterns in our thinking, it is *regulative* of those patterns, articulating an aspiration that needs to be realized by *systematizing* thought—rendering it *more* systematic than it already is.

In addition to the three possible senses we already discerned in the phrase "the systematicity of thought"—(i) systematicity as a symmetry in our thinking, (ii) systematicity as the microstructure of what is thought, and (iii) systematicity as the macrostructure of what is thought—we thus arrive at a fourth and final sense: (iv) systematicity as a regulative ideal for thinking.

By following Kant (1929) in calling it a *regulative* ideal, I mean to underscore that it articulates an ideal that is typically not fully realized, but that nonetheless exerts a pervasive and guiding influence on our thinking—much as theorizing in the natural sciences has been thought (notably by Kant) to be guided by the regulative assumption of the systematic unity of nature, even if our current theories do not yet offer a systematically unified theory of everything.[21] As Emmet (1994, p. 10) puts it, regulative ideals *assign a direction to practice*—and the ideal of systematicity assigns a direction to the practice of thinking.

Of course, the human practice of thinking is far from always fully living up to this ideal. The various inconsistencies and incoherences of human thought and behavior have been extensively documented by social psychology and the psychology of reasoning.[22] Explaining this cognitive "fragmentation" has also become an active area of research in epistemology.[23]

But the fact that human thought does not always fully live up to the ideal of systematicity is precisely what is being acknowledged by referring to it as a *regulative*

---

[21] Some have questioned that they ever will; see notably Neurath (1935, 17) and especially Cartwright (1983, 1999).

[22] For psychological studies to that effect, see notably Garcia-Marques et al. (2015) and Fazio, Rand, and Pennycook (2019). See Mandelbaum and Ripley (2012), Mandelbaum (2014), and Mandelbaum and Quilty-Dunn (2015) for discussions of the philosophical implications of this research.

[23] See the essays in Borgoni, Kindermann, and Onofri (2021), especially Bendaña and Mandelbaum (2021).

*ideal*. As Kant writes, "systematic unity (as mere idea) is only a *projected* unity" (1929, A647/B675). It acts as a *focus imaginarius* (1929, A644/B672), an ideal vanishing point, whose significance lies in the fact that it coordinates cognition even when not fully realized. Just as the ideal vanishing point is called "ideal" because it exists only in the projective extension of the standard Euclidean plane instead of being a tangible part of a real-world scene, so the regulative ideal of systematicity exists only as a projective extension of our actual practice of thinking. What is real, however, is the way this ideal regulates the practice.

Moreover, it is the very fact that human thinking is not always fully systematic that *gives point to* the regulative ideal of systematicity. It provides a corrective, a counterweight, a disciplining normative pull that restrains the proliferation of contradiction, disconnectedness, lacunarity, unprincipledness, or, on the contrary, profligacy of principles. The regulative ideal of systematicity would be pointless if human thought were always fully systematic.

Treating macrosystematicity as a regulative ideal also introduces two complications, however. First, treating macrosystematicity not as a given, but as something that continually needs to be achieved and maintained through systematization, forces one to confront the difficult choices arising from the fact that the five dimensions can be *antagonistic*, i.e. not all fully co-realizable without trade-offs. Comprehensiveness may only be achievable at the expense of parsimony, for example; likewise, articulating general principles in the name of principledness may introduce inconsistencies elsewhere.[24] Just as designing a car involves striking a reasonable balance between speed and safety, the task of systematization involves striking a reasonable balance between various dimensions of systematicity.[25]

Second, treating macrosystematicity as a regulative ideal introduces, as a kind of meta-demand that is prior to the demand to maximize the five dimensions encoded in the 3C2P conception, a demand for *explicitness*: the discursive articulation of what otherwise remains implicit. Before we can even ask to what extent a set of propositions exhibits macrosystematicity, these propositions need to be articulated and discursively laid out. A fundamental part of the task of systematizing ethical thought, for example, consists in transposing the values and intuitions implicit in one's practice into explicitly stateable claims. Only once they are discursively spelled out can we begin to decide which claims articulate fundamental principles, which are applications of those principles, and which should be overturned as contradicting those principles.

This emphasis on the importance of discursively laying out what otherwise remains implicit goes back at least to Socrates, who famously distrusted what could not be discursively spelled out.[26] And it is all the more pronounced in modern liberal democracies, with their ideal of resting authority not in tradition or charisma, but in transparently articulated rules that offer principled justifications for the exercise of public power.[27]

---

[24] See Brun (2020, 950).
[25] See Rescher (1979, 17).
[26] On this aspect of Socrates's thought, see notably Williams (2006).

What the history of the ideal of systematicity reveals is not just that human thinkers have long come to expect some effort at systematization from each other, however. More pertinently for our purposes, it stands to reveal what *drives* systematization.

Understanding what drives systematization is crucial because it is not self-evident that the cognitive ideals we hold each other to should also be extended to AI models. If humans themselves are far from always fully realizing the ideal of systematicity, one might ask, what is the point of measuring the outputs of AI models against that ideal? What is gained by holding AI models to such a standard, and what is lost by their failing to meet it? To answer those questions, I propose to reach back to older discussions of systematicity to recover a sense of the animating *rationales* for the systematization of thought.

## 6 Five Rationales for Systematization

What does cognitive systematization do for us? What are the rationales for it? These are not rationales that the agents performing the systematization need be conscious of; as Dennett (2012) has pointed out, rationales can be "free-floating." Yet reconstructing the rationales for systematization is crucial to determining whether the ideal of systematicity should be applied not just to human intelligence, but to artificial intelligence as well, because it puts us in a position to assess whether—and in what form—the rationales themselves *carry over* from human to artificial intelligence. Our sense of whether opaque AI models should even be held to the ideal of systematic thought is only as clear as our understanding of (a) the rationales for cognitive systematization in the original context of human cognition, and (b) the extent to which these rationales carry over to artificial cognition.

A fundamental rationale for at least some degree of systematization falls out of Enlightenment discussions of the topic. The theory of systematization came into its own with Wolff's *De differentia intellectus systematici & non systematici* (2019), Lambert's fragments on "systematology" (1988), and Kant's "architectonic of pure reason" (1929). And one important insight that emerges from these Enlightenment discussions of systematicity is that *striving* for macrosystematicity is *constitutive* of cognition: were one's perceptions and judgments not subjected to the regulative ideal of systematicity at all, they would not even *purport* to be of determinate objects. Someone who perceived a stick as *rigid and straight* when out of the water and as *rigid and bent* when half-submerged, but remained utterly indifferent to the inconsistency between these two appearances, would not be treating them as two modes of appearance *of one and the same object*.[28] To be completely indifferent to the consistency and coherence of one's perceptions and judgments is to rob them of any determinate content or representational purport, and thus to cease to genuinely *cognize* the world at all. The first and most fundamental rationale for systematization is therefore that it serves a *constitutive* function in cognition.

---

[27] On the idea that this demand for discursive explicitness receives special emphasis in modern liberal democracies, see Weber (2019), Adorno et al. (1989), and Harcourt (manuscript).

[28] For a detailed discussion of this example, see Brandom (2019, 75–80).

This is what led Enlightenment thinkers to cast the imperative to systematization as a demand inherent in reason or rationality itself (Franks, 2005, p. 3).[29] The ideal of systematicity informs the very process whereby the sensory manifold is transformed into the perception of an objective world. Thus, Kant observed: "In accordance with reason's legislative prescriptions, our diverse modes of knowledge must not be permitted to be a mere rhapsody, but must form a system" (1929, A832/B860).

Of course, as the title of Wolff's *De differentia intellectus systematici & non systematici* (2019) reminds us, there is still a difference between cognition that is just systematic enough to merit being called cognition, and cognition that merits being qualified as *systematic* cognition. The point that systematization serves a constitutive function in cognition should accordingly not be taken to rule out the possibility of an *intellectus non systematicus*, an unsystematic intellect. The point is rather that even an unsystematic intellect must already *subject its judgments to the regulative ideal of systematicity* to count as perceiving and thinking about objects at all. Even unsystematic intellects must already strive for systematicity in the minimal sense of being bothered by contradictions between their judgments. Cognition cannot be entirely indifferent to systematicity without ceasing to be cognition.

Of course, idealists from Kant through Fichte to Schelling and Hegel were particularly concerned with the constitution of objective *experience* by cognitive *subjects*, whereas current AI models are widely thought to lack subjective experience. But the relevant rationale for systematization is not tied to conscious experience and can be applied to artificial cognition even if we dissociate it from artificial consciousness, as some have argued we should.[30] Even from a third-personal perspective on a cognitive system, the impression of completely *disjointed cognition* erodes the impression of cognition *tout court*. In what sense does an agent still believe that *p* if the same agent also believes that *not-p*? How inconsistent and incoherent can an individual cognitive perspective get before it ceases to be a cognitive perspective altogether?

Even when we ascribe propositional attitudes to another human being, there is a limit to how fragmented these can become without ceasing to count as propositional attitudes altogether. As Bernard Williams notes, "it is not simply that a person will seem inconsistent or contradictory" if their beliefs "change too often for internal reasons" (2002, p. 191) rather than because the world around them has changed. Eventually, their beliefs will cease to be intelligible to us as *propositional attitudes* at all. They will appear as something else—in the case of human beings, they will perhaps appear as mere "propositional moods" (2002, p. 191), Williams suggests; but in the case of AI models, the effect is more likely to be to disrupt the impression that one is interacting with a form of cognition or intelligence altogether.[31]

---

[29] On Kant's aspiration to cognitive systematization, see Rescher (2000, 64–98), Kitcher (1986), Guyer (2003, 2005), Abela (2006), and Ypi (2021). Other influential Enlightenment discussions include Condillac's *Traité des systèmes* (1749), which distinguished "bad" systems resting on mere hypotheses and speculation (e.g. Pre-Socratic systematizations of nature) from "good" systems resting on experience (e.g. Newton's systematization of celestial mechanics). Similarly, D'Alembert's *Discours préliminaire* (1751) rejected the *esprit de système* in favor of the *esprit systématique*: Procrustean efforts to fit the world into preestablished metaphysical systems were to make way for careful studies of phenomena that distilled systematic principles from them.

[30] See, e.g., Koralus (2023, 35), Frankish (2024), and Müller and Löhr (2025).

The lesson we can draw from Enlightenment discussions of systematization, then, is that subjecting judgments to the imperative to systematization is a central aspect of cognition—indeed, that this imperative to systematization plays a *constitutive role in the attribution of cognition* in the following sense: an LLM need not fully *achieve* systematic integration, but it must at least be *interpretable as striving for it* if the sentences it produces are to be interpretable as expressing propositional attitudes at all.[32] Such striving need not presuppose the presence of conative states, desires, or consciousness on the part of AI systems. To "strive" for systematicity, in this context, means exhibiting robust dispositions to (a) produce outputs that conform to the dimensions of macrosystematicity and (b) detect and correct or mitigate deviations from systematicity when they occur (whether through internal mechanisms or in response to external feedback). This functional understanding aligns with approaches like Dennett's "intentional systems theory" (2009), where goal-directed language is used to explain and predict the behavior of complex systems without attributing conative states, desires, or consciousness to them. An LLM might be interpretable as striving for systematicity much as it might be interpretable as striving for grammaticality: it displays a tendency to conform to the ideal and to amend outputs that fall short.

If an LLM avows both the belief that *p* and the belief that not-*p* while demonstrating a manifest absence of any disposition towards acknowledging and resolving the conflict, however, the effect is to disrupt the impression that one is interacting with a form of cognition or intelligence altogether. This does not mean that models which occasionally produce this effect are non-starters. They might still be valuable tools in lots of ways, and they might still be interpretable as cognitive systems *some* of the time. But to persistently maintain their outputs' interpretability *as manifesting a coherent cognitive perspective*, AI models would have to persistently manifest a general aversion to contradiction and incoherence, thereby striving for systematization to that extent. Ongoing attempts to remedy the respects in which LLMs still fall short of that ideal implicitly acknowledge that point, underscoring that what we want from AI systems is not limited to the interpretability or explainability of individual outputs.

The same holds for ongoing efforts to achieve cross-modal alignment in multimodal models, especially in robotics[33]: to truly apprehend the rational interactions across different modalities, the models need to learn not only to *align* linguistic, visual, acoustic, and tactile representations of the world by treating, say, linguistic and visual representations of apples as similar to each other (as measured by the distance between the corresponding feature vectors in a shared embedding space, for instance)[34]; they also need to treat inconsistencies and incoherences between these representations as *something to be remedied*. That is just what it means for the models to treat these different modalities as representing one and the same world. Cognition, even when artificial, need not fully *achieve* systematic integration, but it must at least *strive* for it if it is to count as cognition.

---

[31] On the attribution of mindedness to AI models, see also Müller and Löhr (2025).
[32] On the notion of an LLM being interpretable as expressing propositional attitudes, see Shanahan (2024b, a) and Chalmers (2025).
[33] See Xu et al. (2023) and Li and Tang (2024).

Second, systematization is fundamental not only to cognizing the world, but also to understanding, because it renders thoughts intelligible in terms of their inferential connections to other thoughts. We do not properly understand a judgment, a belief, or an assertion as long as we cannot see what other thoughts it rationally supports or conflicts with, and what other thoughts it is itself rationally supported or undermined by. We only properly understand it insofar as we can situate it within what Wilfrid Sellars, drawing on Kant, called "the space of reasons" (1997, §36).[35] This goes back to the earlier holistic point against Fodor's semantic atomism. It is constitutive of genuinely *understanding* a thought that one can place it in a web of relationships to other thoughts. One does not genuinely understand the thought that *x* is a *dog,* for example, unless one understands other thoughts that are *inferential consequences* of that thought (e.g. if *x* is a *dog*, this *implies* that it can be kept as a *pet*, and it *excludes* its being a *bird*). More generally, one comes to understand a thought by coming to know how it is systematically integrated into a network of inferential connections. As Brandom (2009, p. 172) argues, understanding is not the sudden incandescence of some "inner lightbulb." Understanding is the mastery of certain practical inferential abilities—it consists in knowing one's way around a certain part of the inferential network. This provides what some in the literature on understanding have referred to as "cognitive control" (Hills, 2015).[36] Consequently, the pursuit of understanding can be a powerful motive for cognitive systematization. We can call this the *hermeneutic* function of systematization.

If this is right, it means that where the interpretability of an AI system's output by end-users is concerned, the demand for a certain degree of systematic integration is not a *separate* demand over and beyond the demand for interpretability, but really encompasses the demand for interpretability *through* systematic integration: it is paradigmatically *by* going some way towards making explicit how a given output can be systematically integrated into a consistent, coherent, comprehensive, and parsimoniously principled body of thoughts that a neural network renders its output interpretable. This does not mean that the AI model itself needs to understand. The point is that for *us* to understand the model's outputs, *we* need to be able to see how these outputs systematically relate to other judgments. And to the extent that the model's outputs themselves make these systematic relationships explicit, that will help us grasp these relationships. If an opaque neural network in science gives us a raw hypothesis without making any effort to show how that hypothesis is consistent with the data, how it is rationally supported by particular observations, and how it can be seen as an application of more general principles of the relevant scientific field, we may not properly understand the hypothesis. The desire to understand the output will then be a powerful motive for wanting the model to make explicit how its output can be systematically integrated with other thoughts. But if we had a model capable of accurately making that systematic integration explicit, it would be for the benefit of our understanding, not the model's. In this sense, the regulative ideal of systematicity is fundamental to making outputs interpretable. The second rationale for systematization, then, is that it serves a hermeneutic function.

---

[34] Xiao et al. (2023), for example, emphasize the need for future research to focus on multimodal interaction.

[35] See also Rescher (2005, 11–13) and Brandom (1994, 2000, 2009).

[36] A congenial emphasis on inferential connections is also found in other discussions of understanding; see notably Kvanvig (2003), Grimm (2006, 2016, 2021), and Baumberger and Brun (2016).

Third, integrability within a system of thoughts acts as a *criterion* for the acceptability of judgments. This provides an *epistemological* rationale for systematization. That point was emphasized by British Neo-Hegelians such as Bradley, who sought to establish "the claim of system as an arbiter of fact" (1909, p. 489). But intimations of this idea can already be gleaned from Plato's insistence, in the *Theaetetus* (201c–210d), that a claim needs a *logos* (a reason) to count as knowledge; or from Aristotle's thesis, in the *Metaphysics* (I, 982a) and throughout the *Posterior Analytics*, that *episteme* is knowledge of certain principles and causes (a thesis that Thomas Aquinas influentially rendered as *sapientis est ordinare*—"the task of the wise is to systematize").[37] More recently, this is also what Nicholas Rescher has extolled as the main purpose of cognitive systematization: it enables "quality control of knowledge claims" (2005, p. 27). Integrability within a system offers a *criterion* by which to separate genuine knowledge from mere opinion.

This has been thought to be one of the principal reasons why academic disciplines strive to be even more systematic than ordinary thought. A high degree of systematicity is widely regarded as a hallmark of *science* in the broad sense of *Wissenschaft*—i.e. the sense in which it covers every discipline taught at a research university. This is the central finding of Paul Hoyningen-Huene's *Systematicity: The Nature of Science* (2013). The claim is not that systematicity is the ultimate test of a scientific theory's acceptability. It is a descriptive claim, to the effect that a movement towards greater systematicity is characteristic of the process whereby a domain of ordinary thought is turned into a *Wissenschaft*. Of course, empiricists like Bas van Fraassen (1980, 2002) would still want to temper this rationalist emphasis by reminding us that systematicity alone is not yet sufficient for scientific success: systematicity may help organize scientific thought, but it does not guarantee truth about unobservables, and the real test of a scientific theory is not its systematicity, but its empirical adequacy.[38] Nonetheless, even van Fraassen accepts Ernest Nagel's

---

[37] See Aquinas (1969, lec. 1, 1).

[38] This ties in with the realism/anti-realism debate in the philosophy of science over whether the theoretical entities postulated by science are "fictions facilitating systematic account" (van Fraassen 1980, 204) or true descriptions of unobservable reality. But note that even an anti-realist like van Fraassen does not

observation that "[i]t is the desire for explanations which are at once systematic and controllable by factual evidence that generates science" (van Fraassen, 1980, p. 92). And before the question of the acceptability of the theory as a whole can even arise, we need to decide which claims to accept into our scientific theory in the first place, and here, systematic integrability does function as a criterion.

The epistemological function of systematic integrability as a criterion for the acceptability of individual claims is crucial to our dealings with AI models, because the basic problem of trust that we face when confronted with their output is, at heart, a problem about whether we should *accept* their output. And if a key criterion of acceptability is whether a claim can be systematically integrated into the wider system of things we take to be true, it is only natural that we should want the outputs of these models to be *systematic* in the sense of being explicitly rationally supported by further explanations or justifications—not just any explanations or justifications, however; a *good* explanation or justification is one that is consistent with other explanations and justifications and coheres with them through relations of rational support; moreover, these explanations and justifications should not seem ad hoc, or multiply to become as numerous as the items they are invoked to support, but should score highly along further dimensions of systematicity, such as comprehensiveness, principledness, and parsimony.

This brings out the important point that getting an AI model to reliably accompany its output with explanations or justifications is not enough. Even highly principled explanations or justifications are not enough. We also want our explanations and justifications to be *unified* as far as possible. This means not only that the principles should, as far as possible, be consistent and coherent, but that we value *economy* of principles. We do not want an endlessly inventive AI that assures us: "If you don't like my principles, I have others."

In other words, we do not just want any kind of explainability; we want explainability *through systematicity*. And once this is recognized, we see that the demand for explainability is really but a facet of a broader demand for systematicity. The demand for explainability can be subsumed under the broader demand for systematicity.

A fourth rationale is that systematization also performs an important *critical* function. Deciding whether to accept a candidate belief into one's existing body of beliefs is not the only form of epistemological quality control we need to engage in. We also need to monitor whether tensions have arisen within the body of beliefs we already possess. And to this end, we need to make the effort to hold up entrenched beliefs to critical scrutiny. The imperative to systematize can instigate such critical revision. Attempting to systematically integrate various judgments can be a way of bringing to light latent tensions between the presuppositions or implications of judgments we take for granted. Systematization in this revisionary role figures prominently in the work of certain ethical theorists such as Ross (1930), Hare (1952, 1972,

---

Footnote 38 (continued)

reject systematization; rather, the anti-realist cautions against mistaking the entities postulated to facilitate systematization for descriptions of unobservable reality.

1989, 2002), and Kagan (1989), who invite us to lean on our principles to uproot our prejudices.[39] The drive towards systematization is an important driver of critique.

In interactions with AI models we do not fully trust, this critical function of systematization can be performed in two directions: an AI model capable of systematization can help one think through the implications of one's own views and awaken one to inconsistencies in them; or the critical leverage generated by systematization can be applied to the AI model itself. By pushing an AI model to explicitly integrate its output into a wider network of thoughts and demonstrating the systematicity of that network, one renders inconsistencies and other flaws glaringly obvious. Systematization then enables others to verify that some judgment or decision is indeed based on the right kind of supporting considerations. This is why people in positions of public authority—such as judges, government commissions, or hospital ethics committees—are expected not just to hand down decisions, but to make discursively explicit how these decisions follow from more general principles that have been consistently and coherently applied (Cueni & Queloz, 2021). That is part of what it means for the decision-making to be fair and to treat like cases alike. Systematicity provides accountability, and enables those at the receiving end of these decisions to *ascertain* their fairness.

Rendering fairness verifiable through systematicity is clearly a desirable feature in AI models. As the fast-growing literature on "fair-AI" and the various techniques for "de-biasing" these models brings out, the anxiety provoked by the opacity of these models stems in part from the worry that they might be biased in ways that are not immediately transparent.[40] Even if de-biasing efforts end up being largely successful, however, those at the receiving end of AI-assisted decisions will still want to *ascertain* that they are indeed not victims of a bias.[41] This concern with transparency and verifiability in the name of fairness is thus a significant part of what fuels calls for explainable AI. But the present argument suggests that this demand can be subsumed under a more general demand for *systematic* AI. The kind of critical oversight that clamors for explainability must be after in this context requires some degree of systematicity rather than mere explainability. To demonstrate the fairness of a decision, not just any explanation will do. The explanation needs to be, more specifically, one that makes explicit how the decision is grounded in more general principles and how these have been consistently and coherently applied to the particular case. An explanation capable of demonstrating the fairness of a decision needs to be a *systematizing* explanation. Once again, we need explainability *through* systematicity.

---

[39] Though this way of seeking critical leverage over our ethical beliefs is contentious even within philosophy and has led a variety of critics to question this identification of greater rationality in ethics with greater systematicity; see Berlin (2013a, 2013b, 2002), MacIntyre (1978, 1988, 2013), Taylor (1985, 1989), Williams (1981, 1985), Stocker (1990), Dancy (1995, 2004), Wolf (1982, 1992, 2007, 2010), Chappell (2015), and Geuss (2020).

[40] For a recent "state-of-the-art" of fair-AI methods and resources, see Alvarez et al. (2024). Räz (2024a) distinguishes different notions of fairness in AI and their weaknesses. See also Müller (2023, 2025).

[41] See Vredenburgh (2022) and Deeks (2019).

Fifth, systematization performs a *didactic* function, facilitating synoptic exposition, persuasion, and retention. This is reflected in the studiedly systematic organization of effective textbooks or slide decks in pedagogical contexts. But it also holds for the structure of speeches, reports, encyclopedia entries, or even ordinary conversations in which one person is asked to introduce someone to a topic they are unfamiliar with. In any context in which a complex body of new information is to be effectively imparted, systematization helps get the information across. This too provides an easily overlooked but important rationale for systematization. By having something laid out in terms of an easily surveyable system, we gain a sense of its structure—in the broad sense of how a complex array of thoughts can be *organized into a systematic order*. This renders the complex array of thoughts much easier to convey and internalize; as schema theory explains, organizing information into structured schemas allows for more efficient encoding and retrieval of complex information (Anderson, 1983; Rumelhart, 1980), and cognitive load theory suggests that the systematic organization of content helps manage cognitive load during complex learning tasks (Sweller et al., 1998).

Moreover, systematization also has a *protreptic* effect, i.e. it renders the array of thoughts more *persuasive* by displaying its inner consistency and coherence. According to the Elaboration Likelihood Model, for example, people are more likely to perceive a message as credible and authoritative if it is organized coherently (Petty & Cacioppo, 1986). This protreptic effect relies on the fact that we have historically come to regard systematicity as a criterion of truth and a hallmark of authoritativeness.

Lastly, systematicity is a powerful *mnemonic* device, rendering a complex array of thoughts much easier to memorize. As George Miller's (1956) classic study showed, for instance, boiling down information to a small number (around seven) of meaningful "chunks"—ideally standing for unifying principles from which more particular information can be derived—significantly enhances people's ability to remember new material. If you want to remember a complex body of information, you will be well served by breaking it down into a manageable number of chunks unified by basic principles.

This didactic rationale for systematization is particularly evident in the history of the notion of cognitive systematicity. When the term "system" begins to be applied to systems *of thought* during the Renaissance, this is purposively done to facilitate synoptic exposition, persuasion, and retention. Of course, the *term* "system" goes back to the ancient Greek terms *systema* and *systematikos*, but these were not initially used to express the *concept* at issue here. *Systema* comes from *syn-histemi*, "to (make to) stand together," and the term was originally applied not to a constellation of judgments exhibiting systematicity, but to flocks of animals, formations of soldiers, or composite political units.[42] And while the Stoics began to use the term in a technical sense to refer to the *systema mundi*,[43] the orderly whole of heaven and earth, they still used it to refer to the systematicity of the world rather than of thought.[44] It was only in the sixteenth century that Protestant theologians, reacting to the confusion about what followers of the new faith were supposed to believe, began to publish synoptic expositions of the new theological doctrines, and referred to them as "systems" (Ritschl, 1906, p. 16). Philosophers soon followed suit with works such as Keckerman's *Systema logicae* (1600) and Timpler's *Metaphysicae Systema Methodicum* (1604). But the rationale for these systematizations of theology, logic, and metaphysics was primarily (though not exclusively) a didactic one.

---

[42] Though Plato also speaks of the Pythagorean idea of a "system" of intervals between notes in the *Philebus* (17d).

The didactic rationale clearly transfers well to AI models. It applies paradigmatically to models optimized for teaching. But it also applies more broadly, to AI models acting as interfaces to the world's knowledge, since these gain an edge over traditional search engines precisely by laying out information in more perspicuous and systematic ways, and by making it possible to ask indefinitely many follow-up questions to prompt a deeper systematic integration of the initial response. It is quickly becoming clear that these models stand to fundamentally change the way we access knowledge. Gone are the days when a textbook or an encyclopedia entry said what it said and not a word more. The promise of these models is that, when suitably prompted, they can offer a deeper systematic integration of the knowledge that humans have uploaded on the internet—and not just by connecting that knowledge merely *associatively*, through hyperlinks, but by *rationally integrating* it, rendering one piece of information interpretable and explainable or justifiable in terms of its inferential connection to other pieces of information that the humans who uploaded the relevant pieces had not yet drawn.

To summarize, systematizing thought can perform five functions:

1. The *constitutive* function, whereby striving to eliminate contradictions and incoherence from one's thoughts is constitutive of cognition.
2. The *hermeneutic function*, whereby thoughts are rendered intelligible in terms of their inferential connections to other thoughts.
3. The *epistemological function*, whereby integrability within a system of thoughts acts as a criterion for the acceptability of new thoughts.
4. The *critical function*, whereby the systematization of thought instigates critical revision.
5. The *didactic function*, whereby the systematization of thought facilitates exposition, persuasion, and retention.

Each of these five functions yields a clear rationale for the systematization of thought that transfers from the original context of human cognition to the novel context of artificial cognition. And in light of these five rationales, it emerges that beneath the clamors for interpretability or explainability in AI models lies a deeper and broader cognitive ideal. From the end-user perspective, it is systematicity that is called for, and this broader demand is fundamental to making outputs interpretable to begin with and underlies our sense of what would make for good explanations. It may be that developers, to understand how to build better models, primarily need outputs—especially outputs that are clearly wrong—to be mathematically interpretable and explainable in terms of the mechanisms that produced them. But end-users have a wider variety of needs. For them, AI models must answer not just to a need for mathematical interpretability or mechanistic explanation, but to a variety of needs that all, in different ways, call for the systematic integration of outputs—be it because they need the models to manifest a unified perspective, or because they need to better understand the outputs, assess their credibility, critique them, or learn and memorize them.

---

[43] See the fragment from Chrysippus in Arnim, *Stoicorum Veterum Fragmenta* (1964, vol. 2, p. 168, l. 15).

[44] This contrast between *objective* and *cognitive* notions of systematicity is emphasized by Rudner (1966, 89), whereas Marchal (1975) highlights their common origin.

## 7 The Hard Systematicity Challenge

This brings into view what we can call the "hard systematicity challenge" to connectionism: the challenge of building neural networks that are sensitive to the structure of thought not merely in the narrow sense of being sensitive to how thoughts are composed of recombinable constituents, but in the broader and more demanding sense of striving towards an integrated body of thought that is consistent, coherent, comprehensive, and parsimoniously principled.

The hard systematicity challenge is one that LLMs have yet to meet. One of "the most disconcerting weaknesses of these models," Cameron Buckner remarks of GPT-3-level models, is their "tendency to meander incoherently in longer conversations and their inability to manifest a coherent individual perspective" (2023, p. 283). At first pass, this tendency is not surprising. The models are not *directly* trained to be more systematic. During pre-training through self-supervised learning (SSL) as well as during supervised fine-tuning (SFT), the explicit training objective of autoregressive models like GPT, Gemini, or Claude is next-token prediction, not systematization.

However, the training objective alone does not tell us all that much about how sensitive LLMs in fact are to systematicity, because it tells us little about *how* the models learn to achieve the training objective (Goldstein & Levinstein, 2024; Herrmann & Levinstein, 2025; Levinstein & Herrmann, 2024). They might conceivably learn to be better next-token predictors *by* forming some sensitivity to the dimensions of systematicity we enumerated. After all, current state-of-the-art models are already a great deal more systematic than GPT-3-level models, and considering whether a token would be a consistent and coherent way to continue a text string certainly seems like a good way of tracking its likelihood of actually being the next token in the training data.

Moreover, while the many inconsistencies and incoherences in the pre-training data lead LLMs to embed inconsistencies and incoherences into their parameters along with more systematic sets of propositions, the resulting lack of systematicity can be substantially mitigated through post-training. The carefully curated and labeled datasets assembled for SFT, for example, are presumably selected notably for their consistency, coherence, comprehensiveness, and parsimonious principledness—or rather, there is some selection *against* input–output pairs that are flagrantly inconsistent, incoherent, incomplete, insufficiently principled, or too profligate in their principles. By being fine-tuned on what are effectively paradigms of localized macrosystematicity *within* an input–output pair, the models are *indirectly* being trained to be more systematic, at least within the scope of their responses.

The same holds for other forms of post-training such as reinforcement learning from human feedback (RLHF). Here also, the training objective is not to be systematic, but to maximize the rating ascribed to the LLM's outputs by a "reward model," which has in turn been trained to reflect the preferences of human annotators. Yet these preferences are apt to penalize a lack of systematicity for all the reasons we considered—we do expect some degree of systematicity from human interlocutors, and there are robust rationales for transferring these expectations to LLMs at least to some extent. As a result, RLHF indirectly incentivizes greater systematicity within a given response.

Even so, systematic integration is something that LLMs still struggle with, especially once the amount of text exceeds their context window (Elazar et al., 2021; Jang & Lukasiewicz, 2023; Kumar & Joshi, 2022; Li et al., 2024; Liu et al., 2024; Maharana et al., 2024). Current LLMs perform poorly on benchmarks testing their long-term memory (Maharana et al. 2024) and episodic memory (Huet et al., 2025), making it harder for them to be consistent and coherent throughout complex narratives. And while an LLM may sound like an erudite human in one conversational thread, the illusion soon breaks down once a different set of prompts in a new thread produces a new web of sentences that lacks coherence with, or even flagrantly contradicts, what it asserted in an earlier thread. Rather like the newspapers as characterized by Karl Kraus, LLMs are still too indifferent to whether what they write today contradicts what they wrote yesterday.

To render models more systematic beyond individual responses, they might need to be fine-tuned *directly* for properties like consistency or coherence. Fine-tuning for consistency within the model's output and fine-tuning for consistency with external databases are being explored, for example (Liu et al., 2024; Zhao et al., 2024; Zhou et al., 2024). Retrieval-Augmented Generation (RAG), which seeks to integrate text generation with the retrieval of information from reliable databases, points to another way in which models can be made more systematic, which is by mirroring the systematicity of the fabric of fact itself. And perhaps LLMs can learn to *leverage* the systematic integration of truths to overcome inconsistencies, incoherences, and lacunae in their training data—at least within domains in which the truth is systematic (Queloz, 2025a). There are also ongoing efforts to use self-consistency mechanisms like verification loops, in which the model checks its own outputs for consistency (Chen et al., 2023; Liang et al., 2024; Wang et al., 2022), or to use latent space planning, in which a high-level outline is generated in latent space before the actual text is filled in (Hao et al., 2024; Zhang et al., 2023).

However, as the next and final section will argue, the demand for systematization itself needs to be regulated by the five rationales for systematization. This is because these five rationales, by clarifying *why* systematization is needed, also provide a guiding sense of *when* and *how* AI models need to strive to systematically integrate their output. And once we adopt a suitably dynamic understanding of the need for systematization, it becomes evident that not all AI models need to be maximally systematic all the time.

## 8 A Dynamic Understanding of the Need for Systematization

If we recognize that the regulative ideal of systematicity answers to a practical need for systematization to discharge certain functions, the question "Should AI be systematic?" takes on a different shape. It no longer looks like a binary yes-or-no question, because we can have *more* or *less* of a need for systematization; and the question no longer appears answerable in the absolute, because the extent to which we have a need for systematization depends on the concrete practical context from which the need arises. The need for systematic AI then appears *scalable* and *context-sensitive*—a need that grows out of, and varies with, AI models' context of application.

Accordingly, we can derive a *dynamic* or *input-dependent* understanding of the need for systematic AI by conceptualizing this need as a function of the following three parameters:

[Which AI model] needs to discharge [which function of systematization] for [which human agents]?

The first parameter registers the fact that whether AI needs to be capable of systematization depends notably on which kind of AI model we are talking about—a network trained to detect breast cancer plays a very different role in human affairs from one trained to predict recidivism, or to play Go, or to teach trigonometry lessons.

The second parameter registers the fact that how much explicit systematization is needed in a given context, and along which dimensions (i.e., consistency, coherence, comprehensiveness, principledness, or parsimony), depends crucially on which functions of systematization need to be discharged in that context. Some functions require a greater degree of systematization than others. For instance, the epistemological function, especially in demanding contexts like scientific research, might require high levels of consistency, coherence, *and* comprehensiveness, whereas the basic constitutive function might only require a minimal threshold of consistency and coherence to maintain interpretability as a cognitive agent at all. Furthermore, different functions prioritize different dimensions: ensuring fairness via the critical function might demand, above all, demonstrable *principledness* and *consistency* in applying those principles across cases. In contrast, facilitating understanding via the hermeneutic function might place a premium on *coherence*—showing how an output rationally connects to other beliefs or data—while the didactic function might additionally emphasize *parsimony* to aid learning and retention. Consequently, the weight we give to these different dimensions should depend on the specific rationale—the function—motivating the need for systematization in the first place.

The third parameter, finally, registers the fact that the need for systematic AI also depends on what kind of human agent is dealing with its output. Is it the AI model's own developers, or is it end-users? Are these end-users school children or math teachers? Defendants or judges? What human agents need from an AI model will vary significantly with the role they occupy.

Depending on which values these three parameters take, the resulting understanding of the need for systematic AI will be more or less demanding. If we ramp up all three parameters to maximum scope, for example, this gives us the following, highly demanding conception of systematicity in AI:

> Every AI model needs to discharge all five functions of systematization for every conceivable human agent.

The need for AI to display systematicity of thought would then be ubiquitous. It would also take a highly demanding form, because the systematicity would have to be such as to fulfill all five functions—and not just for a specific kind of end-user, or for a specific group of people at a certain point in history, but for *every conceivable* human agent. Agents occupying the same end-user role, or a certain sociohistorical perspective, share, in virtue of that fact, certain concepts, capacities, interests, values, intuitions, and background assumptions. But the wider the range of the systematization's addressees, the less shared material there is for the systematization to rely on.

At the other extreme, setting all three parameters to their minimum scope effectively obviates the need for systematic AI:

> No AI model needs to discharge any of the functions of systematization for any human agent.

To understand the need for systematic AI as scalable and context-sensitive is to understand it, first, as lying *on a scale somewhere between* these two extremes; and second, as lying *at a different place* on that scale *depending on context*—in particular, on the value of the three parameters.

My suggestion is that we should treat the functions of systematization as the apex in this triangle of parameters. That is, our sense of *when* and *how* AI models need to systematize should be guided by our sense of the *functions* that need to be discharged in a given context. The degree of systematicity should be proportional to the need for systematization, and the need for systematization reflects the need for certain functions to be discharged.

Thus, if we ask which AI models need to discharge the constitutive function, it is not clear that every AI model is subject to that need to the same degree. If there is a need for an AI model to persistently maintain interpretability as a cognitive system, it comes from *us*—it is *our* need for an AI model to come across as having its own cognitive perspective (instead of coming across as a statistical pattern-matcher or a database technology) that calls for a certain degree of systematization. When using a model as an AI companion, for example, maintaining the impression of mindedness is important to us, and having the model disrupt that impression through incoherent and inconsistent meandering undermines the point of using it as a companion. One recent study even found that "perceiving companion chatbots as more conscious and humanlike correlated with … more pronounced social health benefits" (Guingrich & Graziano, forthcoming). But there are plenty of use cases—when using AI models to code, say, or to provide entertaining trivia—where it is fine for the model to come across as a mere pattern matcher or database technology. In other words, it is the third parameter—*for whom* the models need to systematize—that generates a demand for systematization. At the same time, the *kind* of systematization required to discharge this function will above all foreground consistency, and only require a limited degree of coherence and principledness. The other parameters do not seem called for by the need for cognitive interpretability alone.

Insofar as the hermeneutic function needs to be discharged, by contrast, coherence becomes far more important, since mere consistency is often not yet enough to *understand* something. Some measure of principledness is called for as well, since we often need to understand how something particular instantiates something more general in order to properly understand it.[45] Moreover, parsimony will also matter here, since parsimony in subsuming the particular under principles is what gives understanding its unifying power. As Daniel Wilkenfeld puts it, "understanding is a matter of compressing information about the understood so that it can be mentally useful" (2019, p. 2808), and this cognitive compression expresses a striving towards parsimony. Such a pressure to systematize in the service of understanding will be particularly pronounced where AI models are used in fundamental research, whose defining ambition is to trace the particular to a limited number of highly general principles and demonstrate their comprehensive coverage. But most AI use will serve more modest ambitions.

Similarly, when the function to be discharged is epistemological, coherence is needed in addition to consistency, because consistency alone is typically not yet enough to decide whether to accept a given output. A model's ability to make discursively explicit what rationally supports its output will then also be conducive to discharging that function, as will comprehensiveness, principledness, and parsimony to a greater or lesser degree, depending on whether the users bring ordinary or scientific standards of acceptability to the situation (striving for a particularly high degree of systematicity is, as we saw, a hallmark of science). Yet if we ask *which* AI models must discharge this epistemological function and *for whom*, it will only be those where there is even a question of us accepting the output as true or justified. And this is only the case when human users are in the business of fact-finding at all—which is by no means the only purpose for which people use these models. An AI that dutifully demonstrated how to derive everything it said from first principles would be useless for most creative or conversational purposes. (As philosophers

---

[45] I intentionally do not distinguish further here between semantic, objectual, and explanatory understanding in the way that much of the literature does—see Baumberger, Beisbart, and Brun (2017) for an overview—because, on the inferentialist account of semantic understanding given in Sect. 3, there is a level at which all three forms of understanding operate along the same lines (though that does not preclude their being distinguished *within* that shared framework).

soon discover at cocktail parties, there is such a thing as systematizing to the point of being a bore.)

When it is the critical function that needs to be discharged, however, principledness comes to the fore, since tracing a judgement to more general principles, and searching for inconsistencies either between these principles or in their application, is one of the main ways in which we acquire critical leverage over judgements. As the literature on reflective equilibrium shows, this works in both directions, in that particular judgments might lead one to revise principles just as principles might lead one to revise particular judgments.[46] Moreover, the consistency and coherence of the principles themselves becomes an important consideration in this employment of systematization for critical scrutiny, since the consistent and coherent application of consistent and coherent principles is the mark of non-arbitrary power—and one of the anxieties prompting critical scrutiny of AI models is precisely the worry that their output might be arbitrary.

The same logic applies to cases in which AI models are being used to critically reflect on one's own beliefs and principles. Here too, a comparatively high degree of systematization and principled reasoning will be required to uncover tensions that were buried deep enough not to be obvious from the start. Nonetheless, it would require additional and contested philosophical assumptions for this to lead to the conclusion that AI models should be able to systematize all the way to an axiomatized theory.[47] Absent such assumptions, a reflective equilibrium conception, on which they achieve critical leverage by systematizing some of the way, but not all of the way to a small handful of axioms, offers a less contested model of what kind of systematization we are after for critical purposes.

Finally, the didactic function of systematization reminds us that many AI models will be used not to formulate groundbreaking hypotheses or to critique received opinion, but simply to access and absorb humanity's knowledge. Here, the degree of systematization needed will reflect the depth of the inquirer's curiosity and prior knowledge. But the great strength of these models is that the systematization they provide is itself dynamic: unlike static webpages, which are often, and inevitably, either not systematic enough or too systematic for a given use case, an AI model's output can be tailored, on the fly, to the degree of systematization that a human user needs in a given situation. Having a dynamic understanding of the need for systematization helps us appreciate the dynamic character of the systematization that these models offer as a significant feature.

Thus, by understanding the regulative ideal of the systematicity of thought as answering to practical needs, we can conceive of it as scalable and context-sensitive; and by taking the functions of systematization as a guide to *which* models should systematize, *how* they should systematize, and *for whom*, we get a dynamic understanding of the need for systematicity that yields, not a rigid, one-size-fits-all benchmark for all AI models to meet, but a more nuanced ideal that can be tailored to human needs in particular contexts.

---

[46] See Elgin (1983, 1996, 2017) for epistemological applications of the reflective equilibrium model. For a recent reevaluation, see Beisbart and Brun (2024).

[47] See Queloz (2025b, 169–79, 362–75).

## 9 Conclusion

In the time-honored notion of systematicity, we already have a powerful regulative ideal that effectively underlies our aspiration to make the outputs of AI models interpretable and explainable and our sense of what would count as good explanations, because that ideal has shaped our conception of what it means for thought to be rational, authoritative, and scientific. With the original systematicity challenge having lost much of its challengingness, the hard systematicity challenge that remains is to build AI models that are sensitive to this regulative ideal of systematicity. Aiming for this kind of systematic AI promises to give us a far richer sense of what future AI models ought to be capable of. In the meantime, recognizing that the ideal of systematicity informs our perception of these models can elucidate the respects in which they still strike us as inadequate. For, compared to the narrower notions of interpretability and explainability, the ideal of systematic thought captures more of what it means for artificial intelligence to come across as intelligent at all.

**Funding** Open access funding provided by University of Bern. The work was supported by Schweizerischer Nationalfonds zur Förderung der Wissenschaftlichen Forschung.

## References

Abela, P. (2006). The demands of systematicity: Rational judgment and the structure of nature. In G. Bird (Ed.), *A companion to Kant* (pp. 408–422). Blackwell.

Adorno, T. W., Max H., & Eugen K. (1989). 'Die verwaltete Welt oder: Die Krisis des Individuums'. In *Max Horkheimer, Gesammelte Schriften Bd. 13: Nachgelassene Schriften 1949–72* (pp. 121–142). Fischer

Aizawa, K. (2003). *The systematicity arguments*. Springer.

Alvarez, M. (2010). *Kinds of reasons: An essay in the philosophy of action*. Oxford University Press.

Alvarez, J. M., Colmenarejo, A. B., Elobaid, A., Fabbrizzi, S., Fahimi, M., Ferrara, A., Ghodsi, S., Mougan, C., Papageorgiou, I., Reyero, P., Russo, M., Scott, K. M., State, L., Zhao, X., & Ruggieri, S. (2024). Policy advice and best practices on bias and fairness in AI. *Ethics and Information Technology, 26*(2), 31.

Anderson, J. R. (1983). *The architecture of cognition*. Harvard University Press.

Aquinas, T. (1969). *Sancti Thomae de Aquino Opera Omnia: Sententia Libri Ethicorum* (Vol. 1). Ad Sanctae Sabinae.
Aydede, M. (1997). Language of thought: The connectionist contribution. *Minds and Machines, 7*(1), 57–101.
Baumberger, C., & Brun, G. (2016). Dimensions of objectual understanding. In S. R. Grimm, C. Baumberger, & S. Ammon (Eds.), *Explaining understanding: New perspectives from epistemology and philosophy of science* (pp. 165–189). Routledge.
Baumberger, C., Beisbart, C., & Brun, G. (2017). What Is understanding? An overview of recent debates in epistemology and philosophy of science. In S. R. Grimm, C. Baumberger, & S. Ammon (Eds.), *Explaining understanding: New perspectives from epistemology and philosophy of science* (pp. 1–34). Routledge.
Beisbart, C. forthcoming. Epistemology of artificial intelligence. In E. N. Zalta (Ed.), *The stanford encyclopedia of philosophy*.
Beisbart, C., & Brun, G. (2024). Is there a defensible conception of reflective equilibrium? *Synthese, 203*(79), 1–27.
Benchekroun, O., Rahimi, A., Zhang, Q., & Kodliuk, T. (2020). The need for standardized explainability. Preprint retrieved from arXiv:2010.11273.
Bendaña, J., & Mandelbaum, E. (2021). The fragmentation of belief. In C. Borgoni, D. Kindermann, & A. Onofri (Eds.), *The fragmented mind* (pp. 78–107). Oxford University Press.
Berlin, I. (2002). Two concepts of liberty. In H. Hardy (Ed.), *Liberty* (pp. 166–217). Oxford University Press.
Berlin, I. (2013a). The decline of Utopian ideas in the west. In H. Hardy (Ed.), *The crooked timber of humanity: Chapters in the history of ideas* (pp. 21–50). Princeton University Press.
Berlin, I. (2013b). The pursuit of the ideal. In H. Hardy (Ed.), *The crooked timber of humanity: Chapters in the history of ideas* (pp. 1–20). Princeton University Press.
Borgoni, C., Dirk, K., & Andrea, O. (2021). *The fragmented mind*. Oxford University Press.
Bradley, F. H. (1909). Coherence and contradiction. *Mind, 18*(72), 489–508.
Brandom, R. (1994). *Making it explicit: Reasoning, representing, and discursive commitment*. Harvard University Press.
Brandom, R. (2000). *Articulating reasons*. Harvard University Press.
Brandom, R. (2007). Inferentialism and some of its challenges. *Philosophy and Phenomenological Research, 74*(3), 651–676.
Brandom, R. (2008). *Between saying and doing*. Oxford University Press.
Brandom, R. (2009). *Reason in philosophy: Animating ideas*. Belknap Press.
Brandom, R. (2019). *A spirit of trust: A reading of Hegel's phenomenology*. Harvard University Press.
Brun, G. (2020). Conceptual re-engineering: From explication to reflective equilibrium. *Synthese, 197*(3), 925–954.
Buchholz, O. (2023). A means-end account of explainable artificial intelligence. *Synthese, 202*(2), 33.
Buckner, C. (2023). *From deep learning to rational machines: What the history of philosophy can teach us about the future of artificial intelligence*. Oxford University Press.
Buckner, C., & Garson, J. (2019). Connectionism. In E. N. Zalta (Ed.), *The stanford encyclopedia of philosophy* (Fall 2019). Springer.
Buijsman, S. (2022). Defining explanation and explanatory depth in XAI. *Minds and Machines, 32*(3), 563–584.
Butlin, P. (2023). Sharing our concepts with machines. *Erkenntnis, 88*(7), 3079–3095.
Calvo, P., & Symons, J. (Eds.). (2014). *The architecture of cognition: Rethinking Fodor and Pylyshyn's systematicity challenge*. MIT Press.
Camp, E. (2004). The generality constraint and categorial restrictions. *The Philosophical Quarterly, 54*(215), 209–231.
Cartwright, N. (1983). *How the laws of physics lie*. Oxford University Press.
Cartwright, N. (1999). *The dappled world: A study of the boundaries of science*. Cambridge University Press.
Chalmers, D. J. (2025). Propositional interpretability in artificial intelligence. Preprint retrieved from arXiv:2501.15740
Chappell, S. G. (2015). *Intuition, theory, and anti-theory in ethics*. Oxford University Press.
Chen, X., Aksitov, R., Alon, U., Ren, J., Xiao, K., Yin, P., Prakash, S., Sutton, C., Wang, X., & Zhou, D. (2023). Universal self-consistency for large language model generation. Preprint retrieved from arXiv:2311.17311

Churchland, P. S., & Sejnowski, T. J. (1990). Neural representation and neural computation. *Philosophical Perspectives, 4*, 343–382.

Clark, A. (1991). Systematicity, structured representations and cognitive architecture: A reply to Fodor and Pylyshyn. In T. Horgan & J. Tienson (Eds.), *Connectionism and the Philosophy of Mind* (pp. 198–218). Springer.

Colombo, M. (2023). Concept learning in a probabilistic language-of-thought. How is it possible and what does it presuppose? *Behavioral and Brain Sciences, 46*, Article e271.

Condillac, É. B. D. (1749). *Traité des systèmes*. Neaulme.

Confalonieri, R., Coba, L., Wagner, B., & Besold, T. R. (2021). A historical perspective of explainable artificial intelligence wires. *Data Mining and Knowledge Discovery, 11*(1), e1391.

Cueni, D., & Queloz, M. (2021). Whence the demand for ethical theory? *American Philosophical Quarterly, 58*(2), 135–146.

Cummins, R. (1996). Systematicity. *The Journal of Philosophy, 93*(12), 591–614.

Cummins, R. (2010). *The world in the head*. Oxford University Press.

Cummins, R., Blackmon, J., Byrd, D., Poirier, P., Roth, M., & Schwarz, G. (2001). Systematicity and the cognition of structured domains. *Journal of Philosophy, 98*(4), 167–185.

D'Alembert, JLR. (1751). Discours Préliminaire des Éditeurs'. In D. Diderot & JLR D'Alembert (Eds.), *Encyclopédie, ou Dictionnaire raisonné des sciences, des arts et des métiers, par une Société de Gens de lettres*. Vol. 1, i–xlv. University of Chicago, ARTFL Encyclopédie Project (Autumn 2017 Edition), Robert Morrissey and Glenn Roe (Eds.), Retrieved from http://encyclopedie.uchicago.edu/

Dancy, J. (1995). In defense of thick concepts. *Midwest Studies in Philosophy, 20*(1), 263–279.

Dancy, J. (2004). *Ethics without principles*. Clarendon Press.

Deeks, A. (2019). The judicial demand for explainable artificial intelligence. *Columbia Law Review, 119*(7), 1829–1850.

Dennett, D. (2009). Intentional systems theory. In A. Beckermann, B. P. McLaughlin, & S. Walter (Eds.), *The Oxford handbook of philosophy of mind* (pp. 339–350). Oxford University Press.

Dennett, D. C. (2012). The free floating rationales of evolution. *Rivista di Filosofia, 103*(2), 185–200.

Dickie, I. (2010). The generality of particular thought. *The Philosophical Quarterly, 60*(240), 508–531.

Duque Anton, S. D., Schneider, D., & Schotten, H. D. (2022). On explainability in AI-solutions: A cross-domain survey. *International conference on computer safety, reliability, and security* (pp. 235–246). Springer International Publishing.

Elazar, Y., Kassner, N., Ravfogel, S., Ravichander, A., Hovy, E., Schütze, H., & Goldberg, Y. (2021). Measuring and improving consistency in pretrained language models. *Transactions of the Association for Computational Linguistics, 9*, 1012–1031.

Elgin, C. Z. (1983). *With reference to reference*. Hackett.

Elgin, C. Z. (1996). *Considered judgment*. Princeton University Press.

Elgin, C. Z. (2017). *True enough*. MIT Press.

Elhage, N., Nanda, N., Olsson, C., Henighan, T., Joseph, N., Mann, B., Askell, A., Bai, Y., Chen, A., Conerly, T., DasSarma, N., Drain, D., Ganguli, D., Hatfield-Dodds, Z., Hernandez, D., Jones, A., Kernion, J., Lovitt, L., Ndousse, K., … Olah, C. (2021). A mathematical framework for transformer circuits. *Transformer Circuits Thread., 1*(1), 12.

Eliasmith, C. (2013). *How to build a brain: A neural architecture for biological cognition*. Oxford University Press.

Emmet, D. (1994). *The role of the unrealisable: A study in regulative ideals*. Macmillan.

Ettinger, A., Elgohary, A., Phillips, C. & Resnik, P. (2018). Assessing composition in sentence vector representations. *Proceedings of the 27th International Conference on Computational Linguistics*.

Evans, G. (1982). *The varieties of reference*. In J McDowell (Eds.). Clarendon Press

Fazio, L. K., Rand, D. G., & Pennycook, G. (2019). Repetition increases perceived truth equally for plausible and implausible statements. *Psychonomic Bulletin & Review, 26*(5), 1705–1710.

Fermüller, C. G. (2010). Some critical remarks on incompatibility semantics. In M. Pelis & V. Puncochar (Eds.), *The logica yearbook 2008* (pp. 81–96). College Publications.

Fodor, J. A. (1998). *Concepts: Where cognitive science went wrong*. Clarendon Press.

Fodor, J. (2000). *The mind doesn't work that way: The scope and limits of computational psychology*. MIT Press.

Fodor, J. A. (2003). *Hume variations*. Clarendon Press.

Fodor, J., & Lepore, E. (2002). *The compositionality papers*. Oxford University Press.

Explainability Through Systematicity: The Hard Systematicity…    Page 35 of 39    35Fodor, J., & McLaughlin, B. P. (1990). Connectionism and the problem of systematicity: Why Smolensky's solution doesn't work. *Cognition, 35*(2), 183–205.

Fodor, J. A., & Pylyshyn, Z. W. (1988). Connectionism and cognitive architecture: A critical analysis. *Cognition, 28*(1–2), 3–71.

Frankish, K. (2024). What are large language models doing? In A. Strasser (Ed.), *Anna's AI anthology: How to live with smart machines?* Xenomoi.

Franks, P. W. (2005). *All or nothing: Systematicity, transcendental arguments, and skepticism in German idealism*. Harvard University Press.

Garcia-Marques, T., Silva, R. R., Reber, R., & Unkelbach, C. (2015). Hearing a statement now and believing the opposite later. *Journal of Experimental Social Psychology, 56*, 126–129.

Gaukroger, S. (2020). *Civilization and the culture of science: Science and the shaping of modernity, 1795–1935, civilization and the culture of science*. Oxford University Press.

Geuss, R. (2020). *Who needs a worldview?* Harvard University Press.

Gilpin, L. H., Bau, D., Yuan, B. Z., Bajwa, A., Specter, M., & Kagal ,L. (2018). Explaining explanations: An overview of interpretability of machine learning. In 2018 IEEE 5th International Conference on data science and advanced analytics (DSAA).

Glock, H.-J. (2006). Concepts: Representations or abilities? In E. Di Nucci & C. McHugh (Eds.), *Content, consciousness, and perception: Essays in contemporary philosophy of mind* (pp. 36–61). Cambridge Scholars Press.

Glock, H.-J. (2009a). Concepts, conceptual schemes and grammar. *Philosophia, 37*(4), 653.

Glock, H.-J. (2009b). Concepts: Where subjectivism goes wrong. *Philosophy, 84*(1), 5–29.

Glock, H.-J. (2010a). Concepts: Between the subjective and the objective. In J. Cottingham & P. M. S. Hacker (Eds.), *Mind, method, and morality: Essays in honour of Anthony Kenny* (pp. 306–329). Oxford University Press.

Glock, H.-J. (2010b). Wittgenstein on concepts. In A. Ahmed (Ed.), *Wittgenstein's philosophical investigations: A critical guide* (pp. 88–108). Cambridge University Press.

Glock, H.-J. (2020). Concepts and experience: A non-representationalist perspective. In C. Demmerling & D. Schröder (Eds.), *Concepts in thought, action, and emotion: New essays* (pp. 21–41). Routledge.

Goldfarb, W. (1997). Wittgenstein on fixity of meaning. In W. W. Tait (Ed.), *Early analytic philosophy: Frege, Russell, Wittgenstein* (pp. 75–89). Open Court.

Goldstein, S., & Levinstein, B. A. (2024). Does chatgpt have a mind?. Preprnt retrieved from arXiv: 2407.11015.

Goodman, N. D., Tenenbaum, J. B., & Gerstenberg, T. (2015). Concepts in a probabilistic language of thought. In E. Margolis & S. Laurence (Eds.), *The conceptual mind: New directions in the study of concepts* (pp. 623–654). MIT Press.

Grimm, S. R. (2006). Is understanding a species of knowledge? *The British Journal for the Philosophy of Science, 57*(3), 515–535.

Grimm, S. R. (2016). Understanding and transparency. In C. Baumberger, S. R. Grimm, & S. Ammon (Eds.), *Explaining understanding: New perspectives from epistemology and philosophy of science* (pp. 212–229). Routledge.

Grimm, S. R. (2021). Understanding. In E. N. Zalta (Ed.), *The Stanford encyclopedia of philosophy*. Summer.

Guingrich, R. E., & Graziano, M. S. A. Forthcoming. 'Chatbots as social companions: How people perceive consciousness, human likeness, and social health benefits in machines'. *Oxford intersections: AI in SOCIETY*.

Guyer, P. (2003). Kant on the systematicity of nature: Two puzzles. *History of Philosophy Quarterly, 20*(3), 277–295.

Guyer, P. (2005). *Kant's system of nature and freedom*. Oxford University Press.

Hadley, R. F. (1994). Systematicity in connectionist language learning. *Mind and Language, 9*(3), 247–272.

Hadley, R. F. (1997). Cognition, systematicity and nomic necessity. *Mind and Language, 12*(2), 137–153.

Hadley, R. F. (2004). On the proper treatment of semantic systematicity. *Minds and Machines, 14*(2), 145–172.

Hao, S., Sukhbaatar, S., Su, D., Li, X., Hu, Z., Weston, J., & Tian, Y. (2024). Training large language models to reason in a continuous latent space. Preprint retrieved from arXiv:2412.06769.

Harcourt, E. Manuscript. 'Consequentialism, Moralism, and the "Administered World"'.
Springer

Hare, R. M. (1952). *The language of morals*. Oxford University Press.
Hare, R. M. (1972). *Applications of moral philosophy*. Macmillan.
Hare, R. M. (1989). 'Ethical theory and utilitarianism'. In *Essays in ethical theory* (pp. 212–230) Clarendon Press.
Hare, R. M. (2002). A philosophical autobiography. *Utilitas, 14*(3), 269–305.
Herrmann, D. A., & Levinstein, B. A. (2025). Standards for belief representations in LLMs. *Minds and Machines, 35*(1), 1–25.
Hills, A. (2015). Understanding why. *Noûs, 50*(4), 661–688.
Hoyningen-Huene, P. (2013). *Systematicity: The nature of science*. Oxford University Press.
Hsieh, W., Bi, Z., Jiang, C., Liu, J., Peng, B., Zhang, S., Pan, X., Xu, J., Wang, J., & Chen, K. (2024). A comprehensive guide to explainable AI: from classical models to LLMs. Preprint retrieved from arXiv:2412.00800.
Huet, A., Houidi, Z. B., & Rossi, D. (2025). Episodic memories generation and evaluation benchmark for large language models. Preprint retrieved from arXiv:2501.13121.
Hupkes, D., Dankers, V., Mul, M., & Bruni, E. (2020). Compositionality decomposed: How do neural networks generalise? *Journal of Artificial Intelligence Research, 67*, 757–795.
Jang, M. E., & Lukasiewicz, T. (2023). Consistency analysis of chatgpt. Singapore.
Johnson, K. (2004). On the systematicity of language and thought. *The Journal of Philosophy, 101*(3), 111–139.
Kagan, S. (1989). *The limits of morality*. Oxford University Press.
Kant, I. (1929). *Critique of pure reason*. Macmillan and Co.
Keckermann, B. (1600). *Systema logicae*. Antonius.
Kim, B., Wattenberg, M., Gilmer, J., Cai, C., Wexler, J., and Viegas, F. (2018). Interpretability beyond feature attribution: Quantitative testing with concept activation vectors (tcav). In *International conference on machine learning*.
Kitcher, P. (1986). Projecting the order of nature. In R. Butts (Ed.), *Kant's philosophy of material nature* (pp. 201–235). D. Reidel.
Koralus, P. (2023). *Reason and inquiry: The erotetic theory*. Oxford University Press.
Kozlov, M., & Biever, C. (2023). AI 'Breakthrough': Neural net has human-like ability to generalize language. *Nature, 623*, 16–17.
Krishnan, M. (2020). Against interpretability: A critical examination of the interpretability problem in machine learning. *Philosophy & Technology, 33*(3), 487–502.
Kumar, A., & Joshi, A. (2022). Striking a balance: Alleviating inconsistency in pre-trained models for symmetric classification tasks. Dublin, Ireland.
Kvanvig, J. L. (2003). *The value of knowledge and the pursuit of understanding*. Cambridge University Press.
Lake, B. M., & Baroni, M. (2023). Human-like systematic generalization through a meta-learning neural network. *Nature, 623*(7985), 115–121.
Lambert, J. H. (1988). *Texte zur Systematologie und zur Theorie der wissenschaftlichen Erkenntnis*. In G. Siegwart (Ed.), Meiner.
Levinstein, B. A., & Herrmann, D. A. (2024). Still no lie detector for language models: Probing empirical and conceptual roadblocks. *Philosophical Studies*, 1–27.
Li, S., & Tang, H. (2024). Multimodal alignment and fusion: A survey. Preprint retrieved from arXiv:2411.17040.
Li, T., Zhang, G., Do, Q. D., Yue, X., & Chen, W. (2024). Long-context llms struggle with long in-context learning. Preprint retrieved from arXiv:2404.02060.
Liang, X., Song, S., Zheng, Z., Wang, H., Yu, Q., Li, X., Li, R. H., Wang, Y., Wang, Z., & Xiong, F. (2024). Internal consistency and self-feedback in large language models: A survey. Preprint retrieved from arXiv:2407.14507.
Liu, Y., Guo, Z., Liang, T., Shareghi, E., Vulić, I., & Collier, N. (2024). Measuring, evaluating and improving logical consistency in large language models. Preprint retrieved from arXiv:2410.02205.
MacIntyre, A. C. (1978). *Against the self-images of the age*. University of Notre Dame Press.
MacIntyre, A. C. (1988). *Whose justice? Which rationality?* University of Notre Dame Press.
MacIntyre, A. C. (2013). *After virtue: A study in moral theory* (3rd ed.). Bloomsbury Academic.
Maharana, A., Lee, D. H., Tulyakov, S., Bansal, M., Barbieri, F., & Fang, Y. (2024). Evaluating very long-term conversational memory of llm agents. Preprint retrieved from arXiv:2402.17753.
Mandelbaum, E. (2014). Thinking is believing. *Inquiry, 57*(1), 55–96.

Mandelbaum, E., & Quilty-Dunn, J. (2015). Believing without reason, or: Why liberals shouldn't watch fox news. *The Harvard Review of Philosophy, 22*, 42–52.
Mandelbaum, E., & Ripley, D. (2012). Explaining the abstract/concrete paradoxes in moral psychology: The NBAR hypothesis. *Review of Philosophy and Psychology, 3*(3), 351–368.
Mann, S., Crook, B., Kästner, L., Schomäcker, A., & Speith, T. (2023). Sources of opacity in computer systems: Towards a comprehensive taxonomy. In 2023 IEEE 31st International Requirements Engineering Conference Workshops (REW).
Marchal, J. H. (1975). On the concept of a system. *Philosophy of Science, 42*(4), 448–468.
Marcus, G. F. (2001). *The algebraic mind: Integrating connectionism and cognitive science*. MIT Press.
Matthews, R. J. (1994). Three-concept monte: Explanation, implementation and systematicity. *Synthese, 101*(3), 347–363.
McCulloch, W. S., & Pitts, W. (1943). A logical calculus of the ideas immanent in nervous activity. *The Bulletin of Mathematical Biophysics, 5*(4), 115–133.
McLaughlin, B. P. (1993). The connectionism/classicism battle to win souls. *Philosophical Studies, 71*(2), 163–190.
McLaughlin, B. P. (2009). Systematicity redux. *Synthese, 170*, 251–274.
Miller, G. A. (1956). The magical number seven, plus or minus two: Some limits on our capacity for processing information. *Psychological Review, 63*(2), 81–97.
Millière, R., & Buckner, C. (2024). A philosophical introduction to language models—part ii: The way forward. Preprint retrieved from arXiv:2405.03207.
Minh, D., Wang, H. X., Li, Y. F., & Nguyen, T. N. (2022). Explainable artificial intelligence: A comprehensive review. *Artificial Intelligence Review, 55*(5), 3503–3568.
Müller, V. C. (2023). Ethics of artificial intelligence and robotics. In E. N. Zalta & U. Nodelman (Eds.), *The Stanford encyclopedia of philosophy*. Fall.
Müller, V. C. (2025). Philosophy of AI: A structured overview. In N. A. Smuha (Ed.), *Cambridge handbook of the law, ethics and policy of artificial intelligence*. Cambridge University Press.
Müller, V. C., & Löhr, G. (2025). *Artificial minds*. Cambridge University Press.
Nanda, N. (2024). A comprehensive mechanistic interpretability explainer & glossary." *Mechanistic interpretability*. Retrieved from https://dynalist.io/d/n2ZWtnoYHrU1s4vnFSAQ519J.
Neurath, O. (1935). Einheit der Wissenschaft als Aufgabe. *Erkenntnis, 5*, 16–22.
Peacocke, C. (1992). *A study of concepts*. MIT Press.
Pérez, J., Barceló, P., & Marinkovic, J. (2021). Attention is turing-complete. *Journal of Machine Learning Research, 22*(75), 1–35.
Perler, D. (2004). 'Die Systematizität des Denkens: Zu Ockhams Theorie der mentalen Sprache.' *Philosophisches Jahrbuch, 111*(2), 291–311.
Petty, R. E., & Cacioppo, J. T. (1986). *Communication and persuasion: Central and peripheral routes to attitude change*. Springer.
Phillips, S., & Wilson, W. H. (2016). Systematicity and a categorical theory of cognitive architecture: Universal construction in context. *Frontiers in Psychology, 7*, Article 1139.
Piantadosi, S. T. (2021). The computational origin of representation. *Minds and Machines, 31*(1), 1–58.
Piantadosi, S. T., & Hill, F. (2022). Meaning without reference in large language models. Preprint retrieved from arXiv:2208.02957.
Piantadosi, S. T., Muller, D. C. Y., Rule, J. S., Kaushik, K., Gorenstein, M., Leib, E. R., & Sanford, E. (2024). Why concepts are (probably) vectors. *Trends in Cognitive Sciences*. https://doi.org/10.1016/j.tics.2024.06.011
Press, O., Zhang, M., Min, S., Schmidt, L., Smith, N. A., & Lewis, M. (2023). Measuring and narrowing the compositionality gap in language models. *Findings of the Association for Computational Linguistics: EMNLP, 2023*, 5687–5711.
Queloz, M. (2025a). Can AI rely on the systematicity of truth? The challenge of modelling normative domains. *Philosophy & Technology, 38*(34), 1–27.
Queloz, M. (2025b). *The ethics of conceptualization: Tailoring thought and language to need*. Oxford University Press.
Quilty-Dunn, J., Porot, N., & Mandelbaum, E. (2023). The best game in town: The reemergence of the language-of-thought hypothesis across the cognitive sciences. *Behavioral and Brain Sciences, 46*, Article e261.
Quine, W. V. O. (1951). Two dogmas of empiricism. *Philosophical Review, 60*(1), 20–43.
Räz, T. (2024a). 'Gerrymandering individual fairness.' *Artificial Intelligence, 326*, Article 104035.

Räz, T. (2024b). ML interpretability: Simple isn't easy. *Studies in History and Philosophy of Science, 103*, 159–167.
Räz, T., & Beisbart, C. (2024). The importance of understanding deep learning. *Erkenntnis, 89*(5), 1823–1840.
Rescher, N. (1979). *Cognitive systematization: A systems theoretic approach to a coherentist theory of knowledge*. Blackwell.
Rescher, N. (1981). 'Leibniz and the concept of a system'. In *Leibniz's metaphysics of nature: A group of essays* (pp. 29–41). Springer.
Rescher, N. (2000). *Kant and the reach of reason: Studies in Kant's theory of rational systematization*. Cambridge University Press.
Rescher, N. (2005). *Cognitive harmony: The role of systemic harmony in the constitution of knowledge*. University of Pittsburgh Press.
Rescorla, M. (2024). The language of thought hypothesis. In E. N. Zalta (Ed.), *The Stanford encyclopedia of philosophy* (Summer 2024). Stanford.
Ritschl, O. (1906). *System und systematische Methode in der Geschichte des wissenschaftlichen Sprachgebrauchs und der philosophischen Methodologie*. C. Georgi.
Rizzo, M., Veneri, A., Albarelli, A., Lucchese, C., Nobile, M., & Conati, C. (2023). A theoretical framework for ai models explainability with application in biomedicine. In 2023 IEEE Conference on Computational Intelligence in Bioinformatics and Computational Biology (CIBCB).
Ross, W. D. (1930). *The right and the good*. Clarendon Press.
Rudner, R. S. (1966). *Philosophy of social science*. Prentice-Hall.
Rumelhart, D. E. (1980). Schemata: The building blocks of cognition. In R. J. Spiro, B. C. Bruce, & W. F. Brewer (Eds.), *Theoretical issues in reading comprehension* (pp. 33–58). Lawrence Erlbaum Associates.
Russell, S., & Norvig, P. (2021). *Artificial intelligence: A modern approach*. Pearson.
Ryle, G. (2009). Phenomenology versus the concept of mind. *Critical essays: Collected papers* (Vol. 1, pp. 186–204). Routledge.
Salje, L. (2019). Talking our way to systematicity. *Philosophical Studies, 176*(10), 2563–2588.
Sellars, W. (1958). Counterfactuals, dispositions, and the causal modalities. In H. Feigl, M. Scriven, & G. Maxwell (Eds.), *Minnesota studies in the philosophy of science* (Vol. 2, pp. 225–308). University of Minnesota Press.
Sellars, W. (1997). *Empiricism and the philosophy of mind*. In R Rorty (Eds.). Harvard University Press
Shanahan, M. (2024a). Still "Talking about large language models": Some clarifications. Preprint retrieved from arXiv:2412.10291.
Shanahan, M. (2024b). Talking about large language models. *Communications of the ACM, 67*(2), 68–79.
Siegelmann, H. T., & Sontag, E. D. (1992). On the computational power of neural nets. In Proceedings of the fifth annual workshop on computational learning theory.
Smolensky, P. (1988). The constituent structure of connectionist mental states: A reply to Fodor and Pylyshyn. *The Southern Journal of Philosophy, 26*(S1), 137–161.
Smolensky, P. (1990). Tensor product variable binding and the representation of symbolic structures in connectionist systems. *Artificial Intelligence, 46*(1–2), 159–216.
Smolensky, P., McCoy, R. T., Fernandez, R., Goldrick, M., & Gao, J. (2022). Neurocompositional computing: From the central paradox of cognition to a new generation of AI systems. *AI Magazine, 43*(3), 308–322.
Spenader, J., & Blutner, R. (2007). Compositionality and systematicity. In G. Bouma, I. Krämer, & J. Zwarts (Eds.), *Cognitive foundations of interpretation* (pp. 163–174). Royal Netherlands Academy of Arts and Sciences.
Stocker, M. (1990). *Plural and conflicting values*. Clarendon Press.
Sweller, J., Van Merrienboer, J. J. G., & Paas, F. G. W. C. (1998). Cognitive architecture and instructional design. *Educational Psychology Review, 10*, 251–296.
Taylor, C. (1985). *Philosophy and the human sciences: Philosophical papers* (Vol. II). Cambridge University Press.
Taylor, C. (1989). *Sources of the self*. Harvard University Press.
Timpler, C. (1604). *Metaphysicae Systema Methodicum*. Caesar.
Travis, C. (2015). On constraints of generality. *Proceedings of the Aristotelian Society, 94*(1), 165–188.
Turbanti, G. (2017). *Robert Brandom's normative inferentialism*. John Benjamins.
van Fraassen, B. C. (1980). *The scientific image*. Oxford University Press.
van Fraassen, B. C. (2002). *The empirical stance*. Yale University Press.

van Gelder, T. (1990). Compositionality: A connectionist variation on a classical theme. *Cognitive Science, 14*(3), 355–384.
Vaswani, A., Shazeer, N., Parmar, N., Uszkoreit, J., Jones, L., Gomez, A. N., Kaiser, Ł., & Polosukhin, I. (2017). Attention is all you need. *Advances in Neural Information Processing Systems,* 30.
Verdejo, V. M. (2015). The systematicity challenge to anti-representational dynamicism. *Synthese, 192*(3), 701–722.
von Arnim, H. F. A. (1964). *Stoicorum Veterum Fragmenta*. B. G. Teubner.
Vredenburgh, K. (2022). 'The right to explanation.' *Journal of Political Philosophy, 30*(2), 209–229.
Waismann, F. (1979). *Wittgenstein and the Vienna Circle*. In B McGuinness (Eds.). Blackwell
Wang, X., Wei, J., Schuurmans, D., Le, Q., Chi, E., Narang, S., Chowdhery, A., & Zhou, D. (2022). Self-consistency improves chain of thought reasoning in language models. Preprint retrieved from arXiv:2203.11171.
Weber, M. (2019). *Economy and society: A new translation*. In K. Tribe (eds.). Harvard University Press.
Weiss, G., Goldberg, Y., & Yahav, E. (2018). On the practical computational power of finite precision RNNs for language recognition. Preprint retrieved from arXiv:1805.04908.
Wilkenfeld, D. A. (2019). 'Understanding as compression.' *Philosophical Studies, 176*(10), 2807–2831.
Williams, B. (1981). Conflicts of values. In *Moral luck* (pp. 71–82). Cambridge University Press.
Williams, B. (1985). *Ethics and the limits of philosophy* (Classics). Routledge.
Williams, B. (2002). *Truth and truthfulness: An essay in genealogy*. Princeton University Press.
Williams, B. (2006). The legacy of Greek philosophy. In M. Burnyeat (Ed.), *The sense of the past: Essays in the history of philosophy* (pp. 3–48). Princeton University Press.
Wolf, S. (1982). Moral saints. *The Journal of Philosophy, 79*(8), 419–439.
Wolf, S. (1992). Two levels of pluralism. *Ethics, 102*(4), 785–798.
Wolf, S. (2007). Moral psychology and the unity of the virtues. *Ratio, 20*(2), 145–167.
Wolf, S. (2010). *Meaning in life and why it matters*. Princeton University Press.
Wolff, C. (2019). Über den Unterschied zwischen einem systematischen und einem nicht-systematischen Verstand. In M. Albrecht (Ed.). Meiner.
Xiao, X., Liu, J., Wang, Z., Zhou, Y., Qi, Y., Cheng, Q., He, B., & Jiang, S. (2023). Robot learning in the era of foundation models: A survey. Preprint retrieved from arXiv:2311.14379.
Xu, P., Zhu, X., & Clifton, D. A. (2023). Multimodal learning with transformers: A survey. *IEEE Transactions on Pattern Analysis and Machine Intelligence, 45*(10), 12113–12132.
Ypi, L. (2021). *The architectonic of reason: Purposiveness and systematic unity in Kant's critique of pure reason*. Oxford University Press.
Yu, L., & Ettinger, A. (2020). Assessing phrasal representation and composition in transformers. *Proceedings of the 2020 conference on empirical methods in natural language processing (EMNLP)*.
Zhang, Y., Jiatao, Gu., Zhuofeng, Wu., Zhai, S., Susskind, J., & Jaitly, N. (2023). Planner: Generating diversified paragraph via latent language diffusion model. *Advances in Neural Information Processing Systems, 36*, 80178–80190.
Zhao, Y., Yan, L., Sun, W., Xing, G., Wang, S., Meng, C., Cheng, Z., Ren, Z., & Yin, D. (2024). Improving the robustness of large language models via consistency alignment. Preprint retrieved from arXiv:2403.14221.
Zhou, J., Ghaddar, A., Zhang, G., Ma, L., Hu, Y., Pal, S., Coates, M., Wang, B., Zhang, Y., & Hao, J. (2024). Enhancing logical reasoning in large language models through graph-based synthetic data. Preprint retrieved from arXiv:2409.12437.

**Publisher's Note** Springer Nature remains neutral with regard to jurisdictional claims in published maps and institutional affiliations.